\def\year{2021}\relax
\documentclass[letterpaper]{article} 
\usepackage{aaai21}  
\usepackage{times}  
\usepackage{helvet} 
\usepackage{courier}  
\usepackage[hyphens]{url}  
\usepackage{graphicx} 
\usepackage{natbib}  
\usepackage{caption} 
\usepackage{times}
\usepackage{enumitem}
\usepackage{epsfig}
\usepackage{algorithm}
\usepackage{algorithmic}
\usepackage[caption=false]{subfig}
\usepackage[algo2e]{algorithm2e}
\usepackage{graphicx}
\usepackage{amsmath}
\usepackage{tabularx}
\usepackage{amssymb}
\usepackage{makecell}
\usepackage{enumerate}
\usepackage{booktabs}
\usepackage[usenames, dvipsnames]{color}
\usepackage{array}
\usepackage{makecell}
\usepackage[table]{xcolor}
\usepackage[utf8]{inputenc} 
\usepackage{url}            
\usepackage{booktabs}       
\usepackage{amsfonts} 
\usepackage{xr}
\usepackage{nicefrac}       
\usepackage{microtype}      
\usepackage[switch]{lineno}
\usepackage[pagebackref=true,breaklinks=true,letterpaper=true,colorlinks=true,citecolor=black,bookmarks=false]{hyperref}
\nocopyright
\title{Towards Gender-Neutral Face Descriptors for Mitigating Bias in Face Recognition}
\author {
        Prithviraj Dhar*\textsuperscript{1}, Joshua Gleason*\textsuperscript{2}, Hossein Souri\textsuperscript{1}, Carlos D. Castillo\textsuperscript{1}, Rama Chellappa\textsuperscript{1}\\
}

\affiliations {
    \textsuperscript{\rm 1} Johns Hopkins University \\
    \textsuperscript{\rm 2} University of Maryland, College Park \\
    * Equal contribution\\
    \{pdhar1, hsouri1, carlosdc, rchella4\}@jhu.edu, gleason@terpmail.umd.edu
}

\begin{document}
\maketitle

\begin{abstract}
State-of-the-art deep networks implicitly encode gender information while being trained for face recognition. Gender is often viewed as an important attribute with respect to identifying faces. However, the implicit encoding of gender information in face descriptors has two major issues: (a.) It makes the descriptors susceptible to privacy leakage, i.e. a malicious agent can be trained to predict the face gender from such descriptors. (b.) It appears to contribute to gender bias in face recognition, i.e. we find a significant difference in the recognition accuracy of DCNNs on male and female faces.  Therefore, we present a novel `Adversarial Gender De-biasing algorithm (AGENDA)' to reduce the gender information present in face descriptors obtained from  previously trained face recognition networks. We show that AGENDA significantly reduces gender predictability of face descriptors. Consequently, we are also able to reduce gender bias in face verification while maintaining reasonable recognition performance.
\end{abstract}

\section{Introduction}
\label{sec:intro}
In the past few years, the accuracy of face recognition networks has significantly improved \cite{schroff2015facenet,taigman2014deepface,ranjan2019fast,deng2018arcface,bansal2018deep}. These improvements have led to the usage of face recognition systems in a large number of applications. This has raised concerns about bias against protected categories such as age, gender or race. A recent study performed by NIST \cite{grother2019face} found evidence that characteristics such as gender and ethnicity impact the verification and matching performance of existing algorithms. Similarly, \cite{buolamwini2018gender} showed that most face-based gender classifiers perform better on male faces than female faces.\\ 

Several works \cite{wang2019racial,amini2019uncovering,krishnapriya2020issues,vangara2019characterizing,nagpal2019deep,article,cavazos2019accuracy,georgopoulos2020investigating} have recently analyzed and proposed techniques to mitigate bias against race and skintone in face recognition. However, the issue of gender bias has not been widely explored. It is often assumed that gender imbalance in face recognition training datasets is a cause of gender bias in face recognition. However, \cite{albiero2020does} show that to obtain similar verification performance on male and female faces, we need to find the appropriate percentage of male and female identities (which may not be equal) in the training dataset. Finding such an appropriate mixture is not scalable. 
\cite{albiero2020face} show that minimizing the effect of gendered hairstyles improves the matching scores for genuine female pairs but does not improve that of impostor female pairs.\\
\begin{figure}
\centering
{\includegraphics[width=\linewidth]{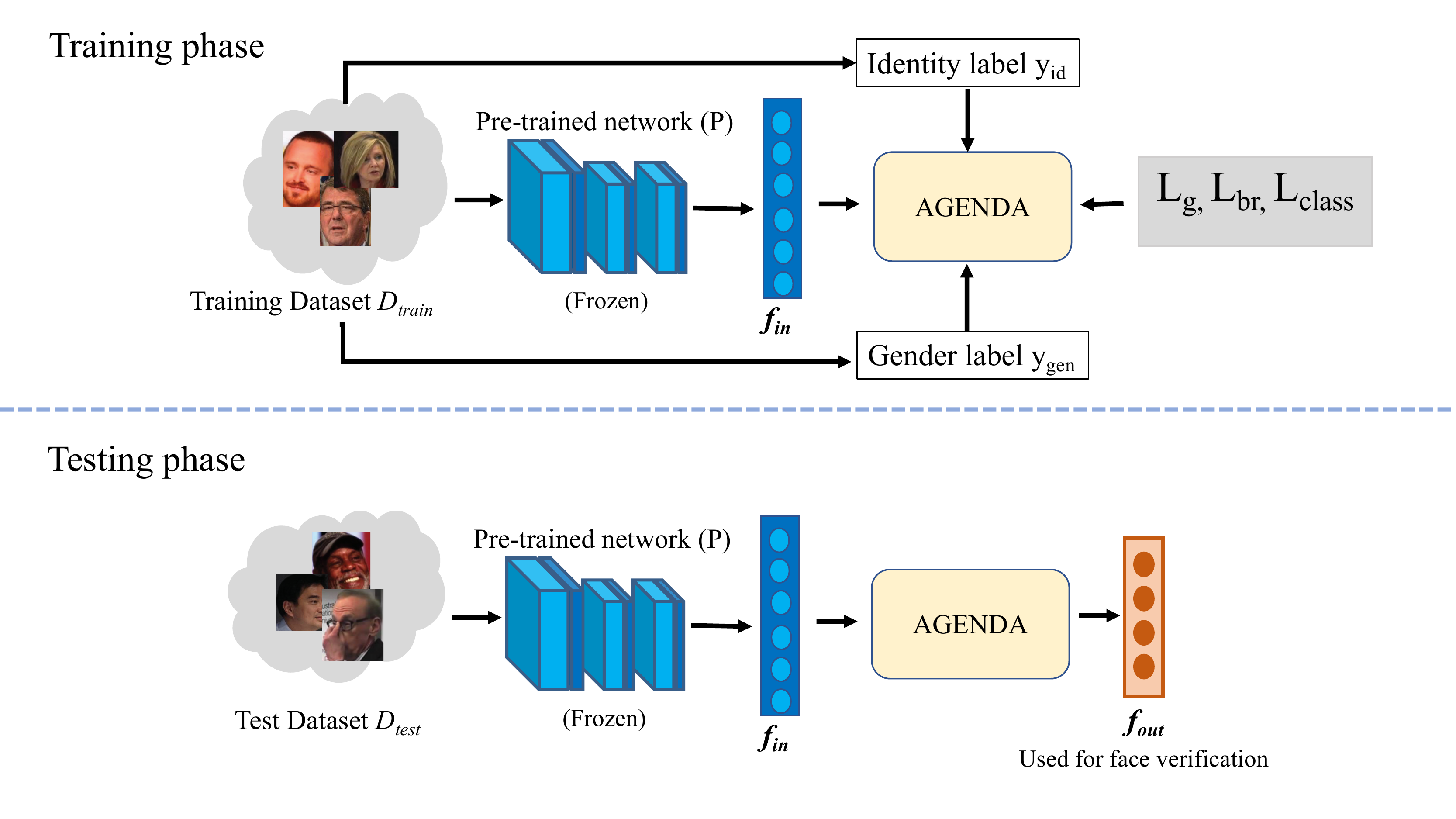}}
\caption{\small We present a framework called AGENDA, which learns to reduce gender information from face descriptors $f_{in}$. After training, we can transform the face descriptors using AGENDA and generate their gender agnostic representations $f_{out}$, which can then be used for face recognition/verification.}
\label{fig:gbteaser}
\vspace{-0.5cm}
\end{figure}

Recent studies \cite{dhar2019attributes, hill2019deep} have also shown that face recognition networks encode gender information while being trained for identity classification. This implies that face descriptors \footnote{Face descriptors refer to the features extracted from the penultimate layer of a previously trained face recognition network.} extracted from such networks can be trained to predict face gender. This can be viewed as privacy leakage since a malicious agent can learn private gender information about an individual without receiving prior authorization. \\

Privacy leakage in face representations is an important issue, as it can allow for the unauthorized extraction of private sensitive attributes (such as race, gender, and age). This issue of estimating soft biometric attributes  in raw facial images has been studied in  \cite{fu2010age,lu2004ethnicity,makinen2008evaluation}. To prevent gender leakage, \cite{othman2014privacy} propose a face-morphing strategy to suppress gender in face images, while preserving identity information. However, there does not exist such a study with respect to post-hoc utilization of face descriptors extracted from previously trained networks. In this regard, we investigate solutions to the following problems: 
\begin{itemize}[leftmargin=3em]
    \item Privacy leakage in face descriptors.
    \item Bias in face verification.
\end{itemize}

%
In our preliminary experiments (in Sec. \ref{subsec:prednbias}), we find that face descriptors from which the gender is more difficult to predict generally demonstrate lower gender bias in face verification tasks. Therefore, we hypothesize that \textit{reducing the ability to predict gender in face descriptors will reduce gender bias in face verification tasks}. To test this hypothesis, we present an \textbf{A}dversarial \textbf{Gen}der \textbf{D}e-biasing \textbf{a}lgorithm (AGENDA) (Fig. \ref{fig:gbteaser}) that transforms face descriptors so that they can be used to accurately classify identity but not gender. Reducing the predictability of gender in face descriptors will impede malicious gender classifiers, thus reducing the possibility of gender leakage. Although in this work we focus on gender leakage and gender bias, with sufficient training data, AGENDA could be modified to reduce the leakage of other sensitive attributes like race and age.


We define gender bias (in Eq. \ref{eq:bias}) as the difference in the face verification performance on male and female faces. As a result of reducing gender predictability (either by using a baseline or AGENDA), we find that gender bias in face verification decreases considerably. Our technique can be used as a post-hoc de-biasing measure for face recognition networks that might exhibit biases. Moreover, our proposed technique does not rely on the gender mixture (i.e. \% of male and female identities) of the training dataset.  To summarize, the contributions of our paper are as follows:
\begin{itemize}[leftmargin=*]
    \item To the best of our knowledge, we are the first to examine verification performance for gender agnostic face descriptors and analyze the subsequent reduction in gender bias.
    \item We experimentally verify that face descriptors with low gender predictability generally demonstrate lower gender bias in face verification. 
    \item To further reduce gender predictability in face descriptors, we propose a method, AGENDA, that unlearns gender information in descriptors while training them for classification. Once trained, AGENDA can then be used to generate gender de-biased representations of face descriptors. When such representations are used for face verification, we find that AGENDA significantly outperforms the baseline with respect to bias reduction. Finally, we analyze the bias versus accuracy trade-off for face verification in male and female faces.
\end{itemize}

\section{Related work}
\textbf{Bias in face recognition:} Several empirical studies \cite{grother2019face, buolamwini2018gender, drozdowski2020demographic} have shown that many publicly available face recognition systems demonstrate bias towards attributes such as race and gender. \cite{wang2019racial, wang2020mitigating} highlight the issue of racial bias in face recognition, and propose strategies to mitigate the same. In the context of gender bias, most experiments show that the face recognition performance on females is lower than that of males. Use of cosmetics by females has been assumed to play a major role in the resulting gender bias \cite{cook2019fixed, klare2012face}. However, \cite{albiero2020analysis} show that cosmetics only play a minor role in the gender gap.\\
\textbf{Building fairer training datasets:} It has been speculated that the unequal percentage of male and female identities in training datasets might lead to gender bias in face recognition. However, \cite{albiero2020does} show that the bias is not mitigated when equal number of male and female identities are used for training. \cite{sattigeri2018fairness} uses a GAN to generate a (Celeb-A \cite{liu2015faceattributes} like) dataset which is less biased with respect to gender, for predicting attractiveness. However, such a method cannot be used to generate unbiased versions of large `in the wild datasets' like MS-Celeb-1M. In contrast to these approaches, our proposed bias mitigation algorithm does not rely on the gender mixture or quality of the training dataset.\\
\textbf{Adversarial techniques to suppress attributes:} \cite{wu2018towards} introduce an approach to anonymize identity and private attributes in a given video, while performing activity recognition. \cite{wang2019balanced} present an adversarial approach to minimize gender leakage in object classification to mitigate bias. Similarly, \cite{alvi2018turning} propose a technique to adversarially minimize the predictability of multiple attributes - gender, pose and ancestral origin, while performing age classification. \cite{li2019deepobfuscator} propose techniques to adversarially minimize gender predictability to reduce gender bias while predicting smile and presence of high-cheekbones. In some of the aforementioned experiments, the attribute under consideration is ephemeral to the target task. For example, in \cite{wu2018towards}, an action is not specific to an identity. Similarly, the presence of smile in \cite{li2019deepobfuscator} may not be unique to a single gender. In contrast, attributes like gender and race may not be ephemeral to face recognition. A given identity can be generally tied to a single gender or race. Therefore, because of the high level of entanglement between identity and gender or race, disentangling them is more involved.\\
\textbf{Gender privacy:} \cite{mirjalili2018gender,mirjalili2017soft} introduce techniques to synthesize perturbed face images using an adversarial approach so that gender classifiers are confounded, but the performance of a commercial face-matcher (in terms of similarity score) is preserved. However, such perturbations have not been demonstrated for pre-trained face descriptors.  It has been already shown in \cite{dhar2019attributes,hill2019deep} that face descriptors implicitly encode gender information during training. Therefore, a  classifier can be easily trained to predict the face gender, using these descriptors as input. Inspired by existing adversarial methods to remove certain attributes, we propose a framework to reduce the gender information in pre-trained face descriptors, while making them efficient for the task of identity classification.  \\

\vspace{-0.5cm}
\section{Problem statement}
Our goal is to reduce gender information in face descriptors so that the ability of a classifier to predict the gender from these descriptors is reduced. From our initial experiments (section \ref{subsec:prednbias}) we find that face descriptors that show low gender predictability demonstrate relatively lower gender bias in face recognition/verification. From this observation, we hypothesize that reducing gender predictability of face descriptors will considerably reduce the gender bias demonstrated by the descriptors. At this point, we quantitatively describe gender bias in the context of face verification.  We define gender bias, at a given false positive rate (FPR) as the absolute difference between the verification performance for male-male and female-female pairs.
\begin{equation}
    \text{Bias}^{(F)} = |\text{TPR}_{m}^{(F)} - \text{TPR}_{f}^{(F)}|
\label{eq:bias} 
\end{equation}
where $\text{TPR}_{m}^{(F)}$ and $\text{TPR}_{f}^{(F)}$ denote the true positive rate for the verification of male-male and female-female pairs respectively at a given FPR $F$. We present a framework that generates de-biased descriptors that, when used for face verification, obtains low bias value and reasonable TPR\textsubscript{f} and TPR\textsubscript{m} at any given FPR.

\begin{figure}
\centering
\hspace*{-0.75cm}\includegraphics[width=1.15\linewidth]{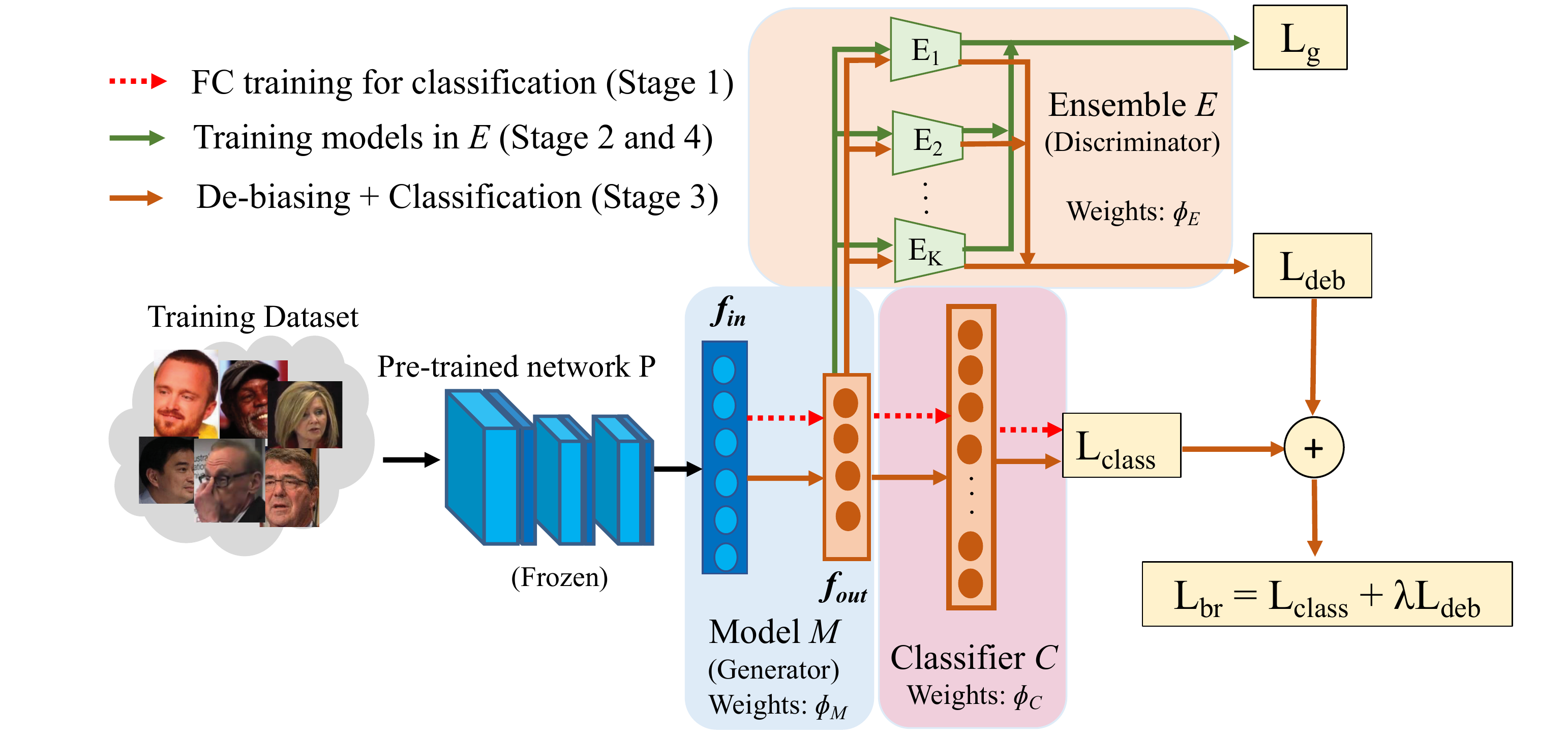}
\caption{\small \textbf{AGENDA architecture}. Face descriptors $f_{in}$ are extracted from a previously trained network $P$ and are fed to a model $M$. $M$ consists of a single linear layer with PReLU \cite{he2015delving} activation that outputs transformed face descriptor $f_{out}$. This is then fed to classifier $C$ and ensemble $E$. The arrows indicate the dataflow at various training stages. In stage 1, $M$ and $C$ are initialized and trained to classify identity using the gradients of $L_{class}$. In stage 2, $E$ is initialized and trained to classify gender using gradients of $L_g$. In stage 3, $M$ and $C$ are trained using the gradients of $L_{br}$ to gender-debias $f_{out}$ while simultaneously being able to classify identity. In stage 4, one member of ensemble $E$ is trained to classify gender from $f_{out}$ using the gradients of $L_g$. Stages 3 and 4 are repeated in alternating fashion, where the ensemble member of $E$ being trained in stage 4 changes at each iteration.}
\label{fig:gbpipeline}
\end{figure}
\section{Proposed approach}
\label{sec:approach}

The key idea in our proposed approach - AGENDA, is to train a model to classify identities while discouraging it to predict gender. Firstly, for a given image $I$, we extract a face descriptor $f_{in}$ using a pre-trained network $P$.
\begin{equation}
    f_{in} = P(I)
\end{equation}
The AGENDA architecture (Fig. \ref{fig:gbpipeline}) is composed of three components:\\
(1) \textbf{Generator model $M$}: A model that takes in face descriptor $f_{in}$ from a pre-trained network $P$, and generates a lower dimensional descriptor $f_{out} \in \mathbb{R}^{256}$. $M$ consists of a single linear layer with 256 units, followed by a PReLU layer. The weights of $M$ are denoted as $\phi_M$. \\
(2) \textbf{Classifier} $C$: A classifier that takes in the output of $M$($f_{out}$) and generates a prediction vector for identity classification. The weights of $C$ are denoted as $\phi_C$.\\
(3) \textbf{Ensemble of discriminators $E$}: An ensemble of $K$ gender prediction models represented as $E_1, E_2 \ldots E_K$. Each of these models is a simple MLP network with an input layer size of 256 units, a SELU activated \cite{klambauer2017self} linear layer with 128 units, and a sigmoid activated output layer with 2 units followed by softmax. We collectively denote the weights of all the models in $E$ as $\phi_E$ and weights of $k^{th}$ model $E_k$ as $\phi_{E_k}$.

We now interpret AGENDA as an adversarial approach. $M$ is a generator that should ideally generate gender agnostic descriptors $f_{out}$. $f_{out}$ is fed to the ensemble $E$ of gender prediction models which acts as a discriminator. $E$ should be able to accurately predict the gender using $f_{out}$. Our aim is to train $M$ to generate descriptors $f_{out}$ that can fool $E$ in terms of gender prediction, and can also be used to classify identities. Therefore, we need to impose two constraints on $f_{out}$: (i.) a penalty term that preserves the identity information, and (ii.) a penalty term that minimizes the gender predictability of $f_{out}$. To this end, we propose a bias reducing classification loss $L_{br}$, which is explained in section~\ref{sec:bias}.
\subsection{Bias reducing classification loss $L_{br}$}
\label{sec:bias}
\setlength{\belowdisplayskip}{1pt}
\setlength{\abovedisplayskip}{1.2pt}
After extracting the descriptor $f_{in}$ from a pre-trained face recognition network, we pass it through $M$ to obtain a lower dimensional descriptor $f_{out}$.
\begin{equation}
    f_{out} = M(f_{in}, \phi_M)
\end{equation}
\textbf{First constraint:} To make $f_{out}$ proficient at classifying identities we provide it to classifier $C$ and use cross-entropy classification loss $L_{class}$ to train both $C$ and $M$.
\begin{equation}
    \mathbf{\hat{y}_{id}} = C(f_{out}, \phi_C)
\end{equation}
\begin{equation}
\label{eq:lclass}
    L_{class}(\phi_M, \phi_C) = -\mathbf{y_{id}} . \text{log}(\mathbf{\hat{y}_{id}})
\end{equation}
$\mathbf{y_{id}}$ is a one hot identity label and $\mathbf{\hat{y}_{id}}$ is the corresponding softmaxed output of classifier $C$.
\\
\textbf{Training discriminators:} $M$ generates $f_{out}$ which is fed to ensemble $E$. Each of the gender prediction models in $E$, denoted $E_k$, are used for computing cross entropy loss $L^{(E_k)}_g$ for gender classification. $L_g$ is computed as the sum of cross-entropy losses for each $E_k$.
\begin{equation}
    L^{(E_{k})}_g(\phi_M, \phi_{E_{k}}) = - y_{g} \text{log } y_{g}^{(k)} - (1-y_{g}) \text{log } (1 - y_{g}^{(k)})
\end{equation}
\begin{equation}
\label{eq:gc}
L_g(\phi_M, \phi_E)=\sum^{K}_{k=1}L^{(E_{k})}_g     
\end{equation}
$y_g$ is the binary gender label for the input face descriptor, and $y^{(k)}_g$ represents the respective softmaxed outputs of $E_k$ in the ensemble.\\ 
\textbf{Training generator (second constraint):} After training $E$, $M$ is trained to transform $f_{in}$ into gender agnostic descriptor $f_{out}$. We then provide $f_{out}$ to each model in ensemble $E$:
\begin{equation}
    o^{male}_{k}, o^{female}_{k} = E_{k}(f_{out}, \phi_{E_{k}})~~\text{for}~ k=1\ldots K
\end{equation}
The outputs $o_k^{*}$ represent the gender probability scores. If an optimal classifier operating on $f_{out}$ were to always produce a posterior probability of 0.5 for both male and female classes then this implies that no gender information is present in the descriptor. To this end, we define the adversarial loss $L^{(E_{k})}_a$ for the $k^{th}$ model in $E$ to be:
\begin{equation}
    L^{(E_{k})}_a(\phi_M,\phi_{E_{k}}) = - (0.5*\text{log}(o^{male}_{k}) + 0.5*\text{log}(o^{female}_{k}))
\end{equation}
Here, we use an ensemble of gender prediction models instead of a single model because we want $f_{out}$ to be gender agnostic with respect to multiple gender predictors. This approach was motivated by the work of \cite{wu2018towards} to solve `the $\forall$ challenge'. After computing the adversarial loss for model $M$ with respect to all the models in $E$, we select the one for which the loss is maximum. We term this loss as debiasing loss $L_{deb}$.
\begin{equation}
\label{eq:ldeb}
    L_{deb}(\phi_M, \phi_E) = \text{max}\{L^{(E_{k})}_a(\phi_M, \phi_{E_k}) |^{K}_{k=1} \}
\end{equation}
The idea is that we would like to penalize $M$ with respect to the strongest gender predictor for which it was not able to fool. This approach was introduced in \cite{wu2018towards}.  $L_{deb}$ is then combined with $L_{class}$ to compute a bias reducing classification loss $L_{br}$ as follows:
\begin{equation}
\label{eq:lbr}
    L_{br}(\phi_C,\phi_M,\phi_E) = L_{class}(\phi_C, \phi_M)+\lambda L_{deb}(\phi_M, \phi_E)
\end{equation}
Here, $\lambda$ is used to weight the de-biasing loss.
\subsection{Stage-wise Training}
\label{sec:stagewise}
We now explain the various stages of training AGENDA. \\
\textbf{Stage 1 - Initializing and training $M$ and $C$}: Using input descriptors $f_{in}$ from a pre-trained network, we train $M$ and $C$ from scratch for $T_{fc}$ iterations using $L_{class}$ (Eq. \ref{eq:lclass}). \\
\textbf{Stage 2 - Initializing and training $E$}: Once $M$ is trained to perform classification, we feed the outputs $f_{out}$ of $M$ to an ensemble $E$ of $K$ gender prediction models. $E$ is trained from scratch to classify gender for $T_{gtrain}$ iterations using $L_g$ (Eq. \ref{eq:gc}). $\phi_M, \phi_C$ remain unchanged in this stage.\\ 
\textbf{Stage 3 - Update model $M$ and classifier $C$}: Here, $M$ is trained to generate descriptors $f_{out}$ that are proficient in classifying identities and are relatively gender-agnostic. $f_{out}$ is fed to the ensemble $E$ and the classifier $C$, the outputs of which result in $L_{deb}$ (Eq. \ref{eq:ldeb}) and $L_{class}$ (Eq. \ref{eq:lclass}) respectively. We combine them to compute $L_{br}$ (Eq. \ref{eq:lbr}) for training $M$ and $C$ for $T_{deb}$ iterations, while $\phi_E$ remains locked.  While computing $L_{br}$, the gradient updates for $L_{deb}$ are propagated to $\phi_M$ and those for $L_{class}$ are propagated to both $\phi_M$ and $\phi_C$.\\
\textbf{Stage 4 - Update ensemble $E$ (discriminator)}: In stage 3, $M$ is trained to generate gender-debiased descriptors $f_{out}$ to fool the models in $E$, whereas in stage 4, models in $E$ are re-trained to classify gender using $f_{out}$. Therefore, we run stages 3 and 4 alternatively, for $T_{ep}$ episodes, after which we re-initialize and re-train all the models in $E$ (as done in stage 2). Here, one episode indicates an instance of running stages 3 and 4 consecutively. In stage 4, we heuristically choose one of the models in $E$, and train it for $T_{plat}$ iterations or until it reaches an accuracy of $G_{thresh}$ on the validation set. $\phi_M$ and $\phi_C$ remain locked in this stage. The motivation for training a single model in $E$ at a time is deferred to discussion in the supplementary material. The full details of training are described in Algorithm \ref{alg:AGENDA}.\\
\vspace{-0.5cm}
\begin{algorithm}[H]
\algsetup{linenosize=\scriptsize}
 \scriptsize
\floatname{algorithm}{Algorithm}
\caption{AGENDA}
\label{alg:AGENDA}
\begin{algorithmic}[1]
\STATE \textbf{Required}: $N_{ep}$: Number of training episodes
\STATE \textbf{Required}: Hyperparams: $\lambda, K, T_{fc}, G_{thresh}, T_{deb}, T_{gtrain}, T_{plat}, T_{ep}$\\
\STATE \textbf{Required} Learning rates: $\alpha_1, \alpha_2, \alpha_3$
\FOR {$i$ in \textbf{range}($N_{ep}$)}
\STATE Begin \textbf{Stage 1} (initial training of $M$ and $C$)
\IF{$i$ == 0}
\STATE Initialize $\phi_M$ and $\phi_C$ with random weights
\FOR {$n$ in \textbf{range}($T_{fc}$)}
\STATE $\phi_M\longleftarrow \phi_M - \alpha_1\nabla_{\phi_M}L_{class}(\phi_M,\phi_C)$
\STATE $\phi_C\longleftarrow \phi_C - \alpha_1\nabla_{\phi_C}L_{class}(\phi_M,\phi_C)$
\ENDFOR
\ENDIF
\STATE Begin \textbf{Stage 2} (initial training of $E$)
\IF {$i$ mod $T_{ep}$ == 0}
\STATE Initialize $\phi_E$ with random weights
\FOR{$n$ in \textbf{range}($T_{gtrain}$)}
\STATE $\phi_E \longleftarrow \phi_E - \alpha_2\nabla_{\phi_E}L_g(\phi_M, \phi_E)$
\ENDFOR
\ENDIF
\STATE Begin \textbf{Stage 3} (update $M$ and $C$)
\FOR{$n$ in \textbf{range}($T_{deb}$)}
\STATE $\phi_M \longleftarrow \phi_M - \alpha_3 \nabla_{\phi_M}L_{br}(\phi_C, \phi_M, \phi_E)$
\STATE $\phi_C \longleftarrow \phi_C - \alpha_3 \nabla_{\phi_C}L_{br}(\phi_C, \phi_M, \phi_E)$
\ENDFOR
\STATE Begin \textbf{Stage 4} (update $E_k$)
\STATE $k$ = $i$ mod $K$
\FOR{$n$ in \textbf{range}($T_{plat}$)}
\STATE Compute validation gender prediction accuracy $A$ of $E_k$
\IF{$A>G_{thresh}$}
\STATE break
\ENDIF
\STATE $\phi_{E_{k}} \longleftarrow \phi_{E_{k}} - \alpha_2 \nabla_{\phi_{E_k}} L^{(E_k)}_{g}(\phi_M, \phi_{E_{k}})$
\ENDFOR
\ENDFOR
\end{algorithmic}
\end{algorithm}
\section{Experiments}
\subsection{Pre-trained networks and evaluation dataset}
We separately evaluate the face descriptors, obtained from the penultimate layer of following two pre-trained networks:\\
\textbf{Arcface} : Resnet-101 trained on MS1MV2 \footnote{https://github.com/deepinsight/insightface/wiki/Dataset-Zoo} with Additive Angular margin (Arcface) loss \cite{deng2018arcface}. There are 59,563 males and 22,499 females in this dataset.\\
\textbf{Crystalface} : Resnet-101 trained on a mixture of UMDFaces\footnote{\label{note2}http://umdfaces.io/}\cite{bansal2017umdfaces}, UMDFaces-Videos\textsuperscript{\ref{note2}}\cite{bansal2017s} and MS-Celeb-1M \cite{guo2016ms}, with crystal loss \cite{ranjan2019fast}. There are 39,712 males and 18,308 females in this dataset.\\
For evaluation, we use the aligned faces in the IJB-C dataset, and follow the 1:1 face verification protocol defined in \cite{maze2018iarpa}. The alignment is done using \cite{ranjan2017all}. However, instead of verifying all the pairs, we only verify male-male and female-female pairs. There are 6.4 million male-male and 2.1 million female-female pairs defined in the protocol. The gender labels are provided in the dataset. Using Arcface and Crystalface we extract 512 dimensional face descriptors for the aligned faces in the IJB-C dataset which are then used for verification of male-male and female-female pairs. The gender-wise verification plots for Crystalface and Arcface descriptors are provided in Fig. \ref{fig:predcomp}.
\subsection{Relation between gender predictability and bias}
\label{subsec:prednbias}
\begin{figure}
\centering
{\includegraphics[width=1\linewidth]{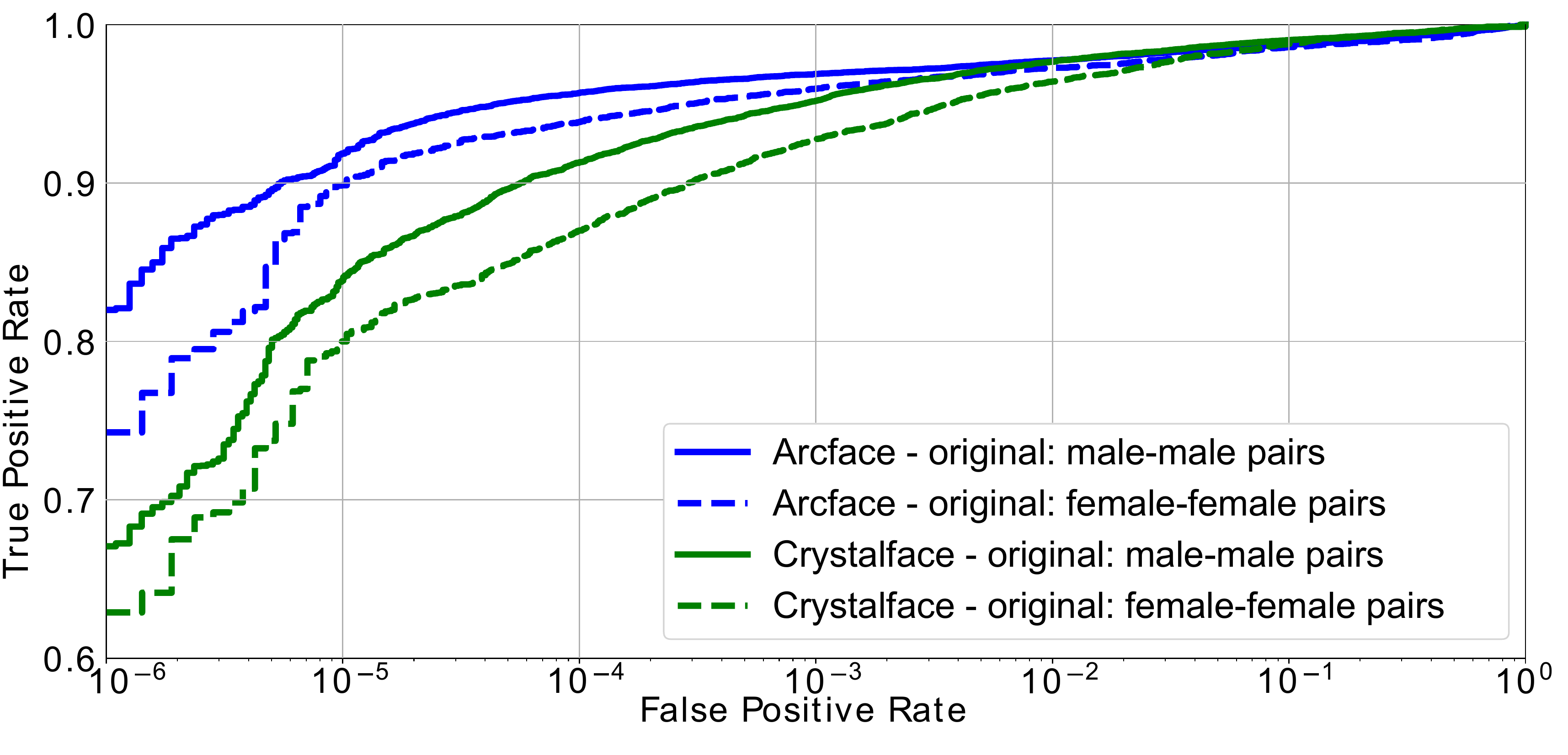}}
\caption{\small Gender-wise verification plots for Arcface and Crystalface}
\vspace{-0.4cm}
\label{fig:predcomp}
\end{figure}

We compare the predictability (i.e. ability to classify gender) of face descriptors extracted from Arcface and Crystalface networks. We train a logistic regression classifier on 60k IJB-C face descriptors (30k males and females) to classify gender and test it on 20k IJB-C descriptors (10k males and females). The images for training and testing are selected randomly, and the face descriptors are extracted using the pre-trained networks (Arcface or Crystalface). Note that there is no overlap between identities in train and test split. From the  results in Table \ref{tab:predcomp}, we find that descriptors extracted using Arcface show relatively low gender predictability than Crystalface descriptors. In Table \ref{tab:predcomp}, we also find that the gender bias (computed using Eq. \ref{eq:bias} and plots shown in Fig \ref{fig:predcomp}) is lower in most FPRs when Arcface descriptors are used for face verification, as comapred to Crystalface descriptors. This shows that \textit{face descriptors with low gender predictability appear to demonstrate lower gender bias in face verification}, thus forming the basis of our initial hypothesis (mentioned in Sec. \ref{sec:intro}). Therefore, we propose techniques to reduce the predictability of gender in face descriptors while making them proficient in identity classification.
\begin{table}
\caption{\small Gender-wise verification results and corresponding bias. Acc. refers to gender prediction accuracy of the logistic regression classifier}
\scriptsize
\centering
\hskip-0.38cm\begin{tabular}{cc|ccc|ccc|ccc|ccc}
\toprule
FPR  & &\multicolumn{3}{c|}{$10^{-6}$}& \multicolumn{3}{c|}{$10^{-5}$} & \multicolumn{3}{c|}{$10^{-4}$} & \multicolumn{3}{c}{$10^{-3}$} \\
\midrule
Network&Acc.& TPR\textsubscript{m} & TPR\textsubscript{f}& Bias & TPR\textsubscript{m} & TPR\textsubscript{f}& Bias & TPR\textsubscript{m} & TPR\textsubscript{f}& Bias & TPR\textsubscript{m} & TPR\textsubscript{f}& Bias\\
\midrule
Arcface& \textbf{76.01}& 0.82&0.74& 0.08 & 0.92 & 0.90 & \textbf{0.02} &0.96 & 0.93 & \textbf{0.03} &0.97&0.96&\textbf{0.01} \\
Crystalface& 80.50& 0.67 &0.63 & \textbf{0.04} & 0.84 &0.8 & 0.04 & 0.92& 0.87 & 0.05&0.96&0.93&0.03\\
\bottomrule
\end{tabular}
 \label{tab:predcomp}
\end{table}
\subsection{Baseline: CorrPCA}
\label{sec:exp1}
 Since our hypothesis involves removing gender specific information from face descriptors, we propose a naive approach, termed as `Correlation-based PCA' (CorrPCA) for this task. We first compute the eigenspace of the descriptors and isolate the eigenvectors that encode gender information. After this, we remove these eigenvectors and transform the test face descriptors using the remaining subspace. A similar approach was used in the early nineties \cite{turk1991face} to reduce the impact of illumination on PCA features.\\ \textbf{Isolating gender specific components:} We first randomly select 80k (40k males and females) aligned face images from MS-Celeb-1M dataset. Then, we extract the 512-dimensional descriptors for these images, using a given pre-trained network (Arcface/Crystalface). We then compute the eigenspace $S\in \mathbb{R}^{512\times512}$ of the descriptors $X \in \mathbb{R}^{80k \times 512}$ using PCA. Using each eigenvector in $S$, we transfrom the original descriptors $X$ as follows :
 \begin{equation}
     v_s = X\cdot s 
 \end{equation}
  Here, $s$ is a row in $S$ (i.e. an eigenvector of $X$), and $v_s$ is the corresponding component of each descriptor. Now, for all the 80k images, we  have a vector $v_s \in \mathbb{R}^{80k\times1}$ and a label vector $\ell$, with gender labels for all the sampled images. After this, we compute the Spearmann correlation coefficient between $v_s$ and $\ell$. We select the eigenvectors in $S$ for which this correlation is lower than $\delta$ and denote them collectively as a subspace $R$. Thus, $R$ is a subspace that has relatively lower gender information. We find that the number of eigenvectors in the subspace $R$ is 504 and 487 for the descriptors of Arcface and Crystalface respectively.\\
 \textbf{Transforming test face descriptors using remaining components:} We then extract the 512-dimensional descriptors for the IJB-C dataset using the given pre-trained network (Arcface/Crystalface), to obtain the transformed descriptors $X_{ijbc}$. Finally, we transform $X_{ijbc}$ using the subspace spanned by $R$ into a new feature space $T_{ijbc}$. We use $\delta=0.1$ in this protocol, for both networks. We evaluate the gender leakage and gender bias of the transformed descriptor sets for IJB-C dataset in Sec. \ref{subsec:res}.\\
\vspace{-0.2cm}
\subsection{Training details for AGENDA}
\label{sec:agendahp}
For training AGENDA, we use a combination of UMDFaces, UMDFaces-Videos and MS-Celeb-1M datasets. The face alignment and gender labels are obtained using \cite{ranjan2017all}. We perform our experiments using input face descriptors $f_{in}$ from Arcface and Crystalface, extracted for these datasets. The hyperparameter information for AGENDA is provided below:\\
\textbf{Stage 1}: $T_{fc}=66000$ iterations, learning rate $\alpha_1 = 10^{-5}$. \\
\textbf{Stage 2}: $T_{gtrain}=30000$ iterations, learning rate $\alpha_2 = 10^{-3}$. Here, we use $K=1$ and $5$ for Arcface and Crystalface, respectively.\\
\textbf{Stage 3}: $T_{deb} = 1200$ iterations,learning rate $\alpha_3 = 10^{-4}$. We compute $L_{br}$  using $\lambda=10$ and $1$, when using $f_{in}$ from Arcface and Crystalface respectively.\\
\textbf{Stage 4}: $T_{plat}=2000$ iterations, learning rate $\alpha_2 = 10^{-3}$ (same as stage 2). $G_{thresh}=0.90$ and $0.80$ for Arcface and Crystalface, respectively.\\
In all the aforementioned training stages, we use an Adam optimizer and a batch size of 400, and we ensure that each batch is balanced in terms of gender.
\begin{figure}
{\centering
\subfloat[]{\includegraphics[height = 0.39\linewidth]{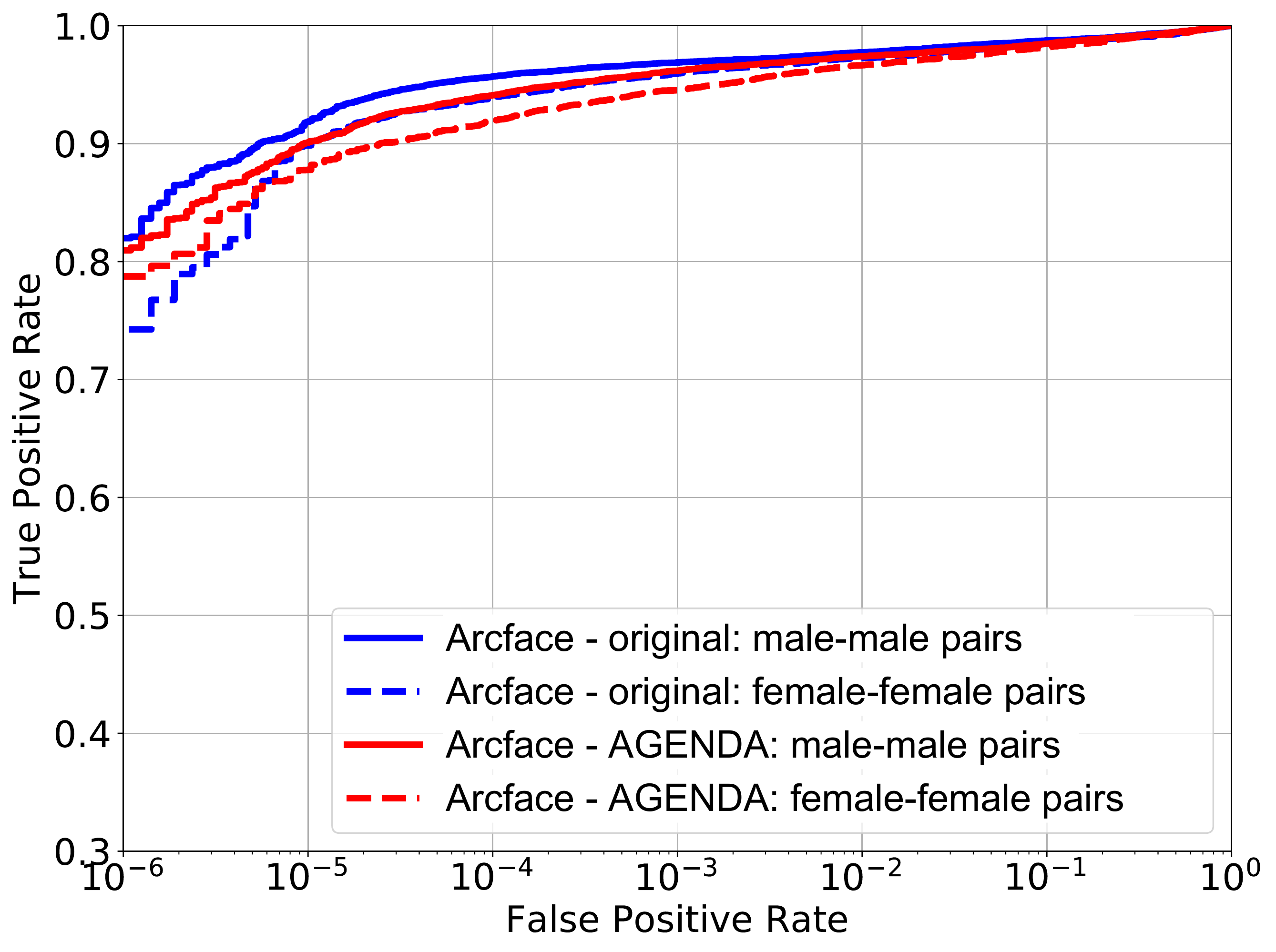}}
\subfloat[]{\includegraphics[height = 0.39\linewidth]{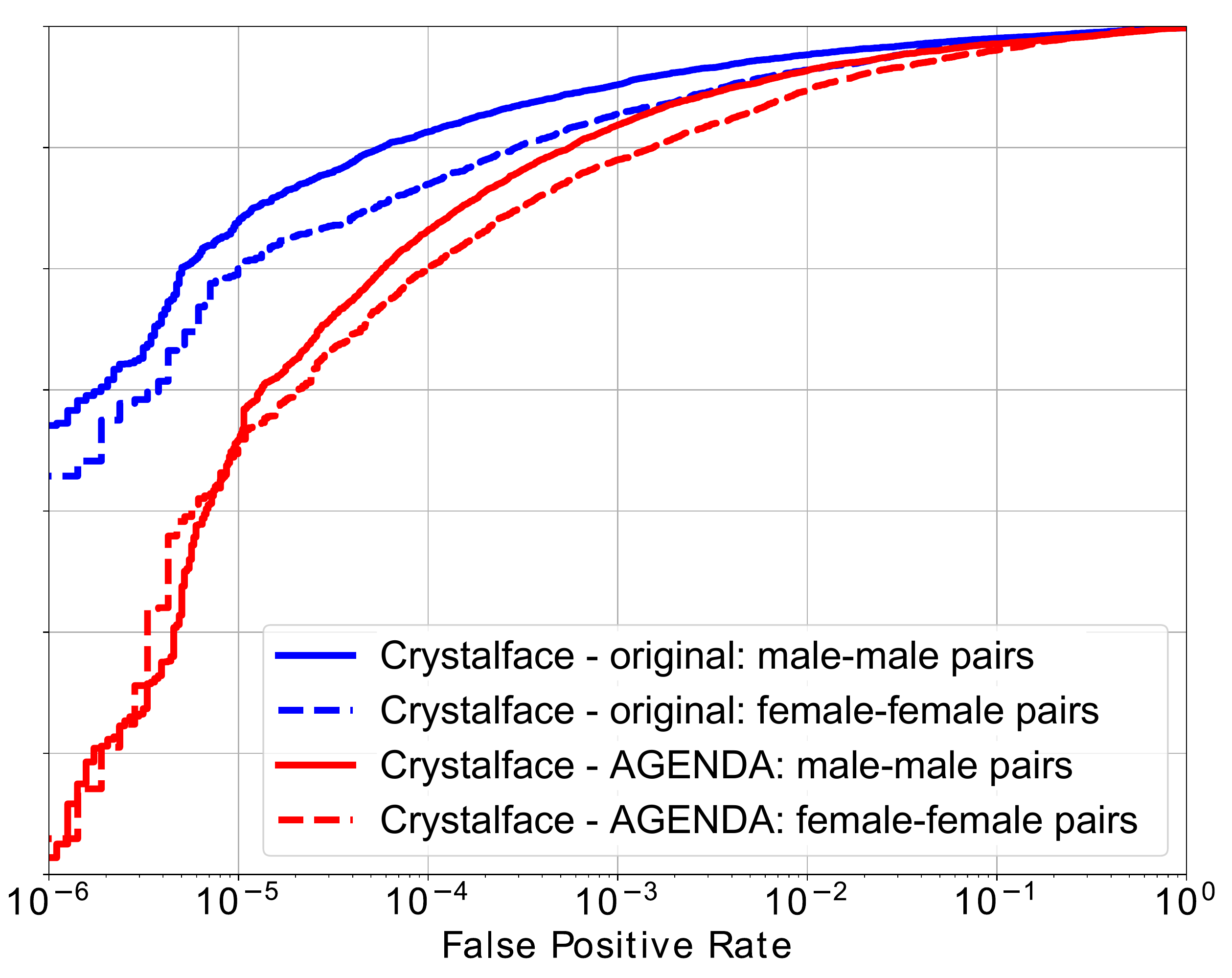}}
\caption{\small Gender-wise IJB-C verification results for (a.) Arcface and (b.) Crystalface using face descriptors from original pre-trained networks and their AGENDA counterparts. AGENDA reduces gender bias at low FPRs.}
\vspace{-0.4cm}
\label{fig:bias_AGENDA}
}
\end{figure}

\subsection{Results}
\label{subsec:res}
\textbf{Evaluating gender leakage}: We follow the same steps as in section ~\ref{subsec:prednbias} for computing the gender classification accuracy (Table \ref{tab:predcomp}) of raw face descriptors. We present the accuracy of the logistic regression classifier trained on face descriptors obtained from CorrPCA and AGENDA in Table \ref{tab:logacc}. We find that for both Arcface, the Crystalface, the classification accuracy goes down when the face descriptors are transformed using AGENDA or CorrPCA framework, which indicates that gender leakage from the descriptors is likely reduced.\\
\textbf{Evaluating verification performance and bias}: We first evaluate the 1:1 verification performance obtained by CorrPCA-transformed IJB-C face descriptors. From Table \ref{tab:biasres}, we can infer that this method helps to reduce the gender bias in Crystalface, whereas the bias and performance in Arcface remains mostly unchanged. Following that, we evaluate the verification performance and corresponding bias obtained after using AGENDA. After AGENDA training, we feed the 512 dimensional $f_{in}$ (extracted from Arcface/Crystalface) of aligned IJB-C images to the trained model $M$ (Fig. \ref{fig:gbpipeline}), that generates 256 dimensional gender de-biased descriptors $f_{out}$. We then use $f_{out}$ to perform gender-wise IJB-C 1:1 face verification. From Fig. \ref{fig:bias_AGENDA}. We find that when using $f_{out}$ from Arcface, the bias is reduced (especially in low FPRs), without catastrophically losing verification performance. This is done by improving female-female verification at low FPRs, while slightly decreasing male-male verification. Similarly, for $f_{out}$ from Crystalface, we find that the gender bias is reduced at low FPRs. The bias is especially close to 0 after FPR $10^{-5}$.\\
From Table \ref{tab:biasres},  we can infer that AGENDA consistently outperforms CorrPCA in terms of bias reduction at almost all the FPRs under consideration.  For FPRs not reported in Table \ref{tab:biasres}, the gender bias in verification is exactly same for CorrPCA, AGENDA and the original pre-trained network.\\
\newcommand{\tabarc}{
\begin{tabular}{cccc|ccc|ccc|ccc}
\toprule
FPR & \multicolumn{3}{c|}{$10^{-6}$}& \multicolumn{3}{c|}{$10^{-5}$} & \multicolumn{3}{c|}{$10^{-4}$} & \multicolumn{3}{c}{$10^{-3}$}\\
\midrule
Method& TPR\textsubscript{m} & TPR\textsubscript{f}& Bias & TPR\textsubscript{m} & TPR\textsubscript{f}& Bias & TPR\textsubscript{m} & TPR\textsubscript{f}& Bias & TPR\textsubscript{m} & TPR\textsubscript{f}& Bias \\
\midrule
Original& 0.82&0.74& 0.08 & 0.92 & 0.90 & 0.02 &0.96 & 0.93 & 0.03 &0.97&0.96&0.01 \\
CorrPCA& 0.82& 0.76&\cellcolor{yellow} 0.06 &0.92 & 0.90 & 0.02 &0.96 & 0.93 & 0.03&0.97&0.96&0.01\\
AGENDA & 0.81&0.79 & \cellcolor{green}\textbf{0.02}&0.90& 0.89& \cellcolor{green}\textbf{0.01} & 0.94 & 0.93 & \cellcolor{green}\textbf{0.01} &0.96&0.95&0.01\\
\bottomrule
\end{tabular}}
\newcommand{\tabrg}{
\begin{tabular}{cccc|ccc|ccc|ccc}
\toprule
FPR & \multicolumn{3}{c|}{$10^{-6}$}& \multicolumn{3}{c|}{$10^{-5}$} & \multicolumn{3}{c|}{$10^{-4}$} & \multicolumn{3}{c}{$10^{-3}$}\\
\midrule
Method & TPR\textsubscript{m} & TPR\textsubscript{f}& Bias & TPR\textsubscript{m} & TPR\textsubscript{f}& Bias & TPR\textsubscript{m} & TPR\textsubscript{f}& Bias & TPR\textsubscript{m} & TPR\textsubscript{f}& Bias\\
\midrule
Original & 0.67 &0.63 & 0.04 & 0.84 &0.8 & 0.04 & 0.92& 0.87 & 0.05&0.96&0.93&0.03\\
CorrPCA & 0.67&0.65&\cellcolor{yellow}0.02&0.84&0.81&\cellcolor{yellow}0.03&0.91&0.87&\cellcolor{yellow}0.04&0.95&0.93&\cellcolor{green}0.02 \\
AGENDA &0.32&0.32& \cellcolor{green}\textbf{0.0}&0.67&0.67&\cellcolor{green}\textbf{0.0}&0.83&0.81&\cellcolor{green}\textbf{0.02}&0.92&0.90&\cellcolor{green}\textbf{0.02} \\
\bottomrule
\end{tabular}}
\newcommand{\tabrgtpe}{
\begin{tabular}{cccc|ccc|ccc|ccc}

\toprule
FPR & \multicolumn{3}{c|}{$10^{-6}$}& \multicolumn{3}{c|}{$10^{-5}$} & \multicolumn{3}{c|}{$10^{-4}$} & \multicolumn{3}{c| }{$10^{-3}$}\\
\midrule
Method & TPR\textsubscript{m} & TPR\textsubscript{f}& Bias & TPR\textsubscript{m} & TPR\textsubscript{f}& Bias & TPR\textsubscript{m} & TPR\textsubscript{f}& Bias & TPR\textsubscript{m} & TPR\textsubscript{f}& Bias\\
\midrule
Original & 0.80&0.69& 0.11 & 0.88 & 0.84 & 0.04 &0.93 & 0.89 & 0.04 &0.96&0.94&0.02\\
CorrPCA & 0.79&0.69&\cellcolor{yellow}0.10&0.88&0.84&0.04&0.93&0.90&\cellcolor{yellow}0.02&0.96&0.94&0.02 \\
AGENDA &0.68&0.60& \cellcolor{green}\textbf{0.08}&0.82&0.80&\cellcolor{green}\textbf{0.02}&0.90&0.87&\cellcolor{green}\textbf{0.03}&0.95&0.93&\cellcolor{green}\textbf{0.02} \\

\bottomrule
\end{tabular}}
\begin{table}
\caption{\small Performance of logistic regression classifier trained using descriptors extracted from original networks, transformed using CorrPCA and extracted using AGENDA.}
\scriptsize
\centering
\begin{tabular}{c|ccc|ccc}
Network & \multicolumn{3}{c|}{Arcface} &  \multicolumn{3}{c}{Crystalface}\\
\hline
 Method & Original & CorrPCA & AGENDA& Original & CorrPCA & AGENDA \\ 
\hline
Gender classif\textsuperscript{n} acc. & 76.01 & 72.33 &64.01 & 80.50 & 75.74 & 67.25 \\
Females misclassified (\%) & 39.15 & 36.08 & 51.10 & 36.63 & 37.70 & 43.82 \\
Males miscalssified (\%) & 8.83 & 20.68 & 20.88 & 4.3& 10.8 & 23.31\\
\end{tabular}
 \label{tab:logacc}
\end{table} 
\begin{table}%
 \caption{\small Gender-wise IJB-C verification results and their corresponding gender bias for (a.) Arcface and (b.) Crystalface. AGENDA shows lower bias than CorrPCA at most FPRs. Green: least bias, Yellow: second least bias.}%
  \scriptsize
  \centering
  \subfloat[][Arcface]{\tabarc}%
  \qquad
  \subfloat[][Crystalface]{\tabrg}
  \label{tab:biasres}%
  \vspace{-0.6cm}
\end{table}

\begin{figure*}
{\centering
\subfloat[Arcface, $\lambda=10$]{\includegraphics[width=0.25\linewidth]{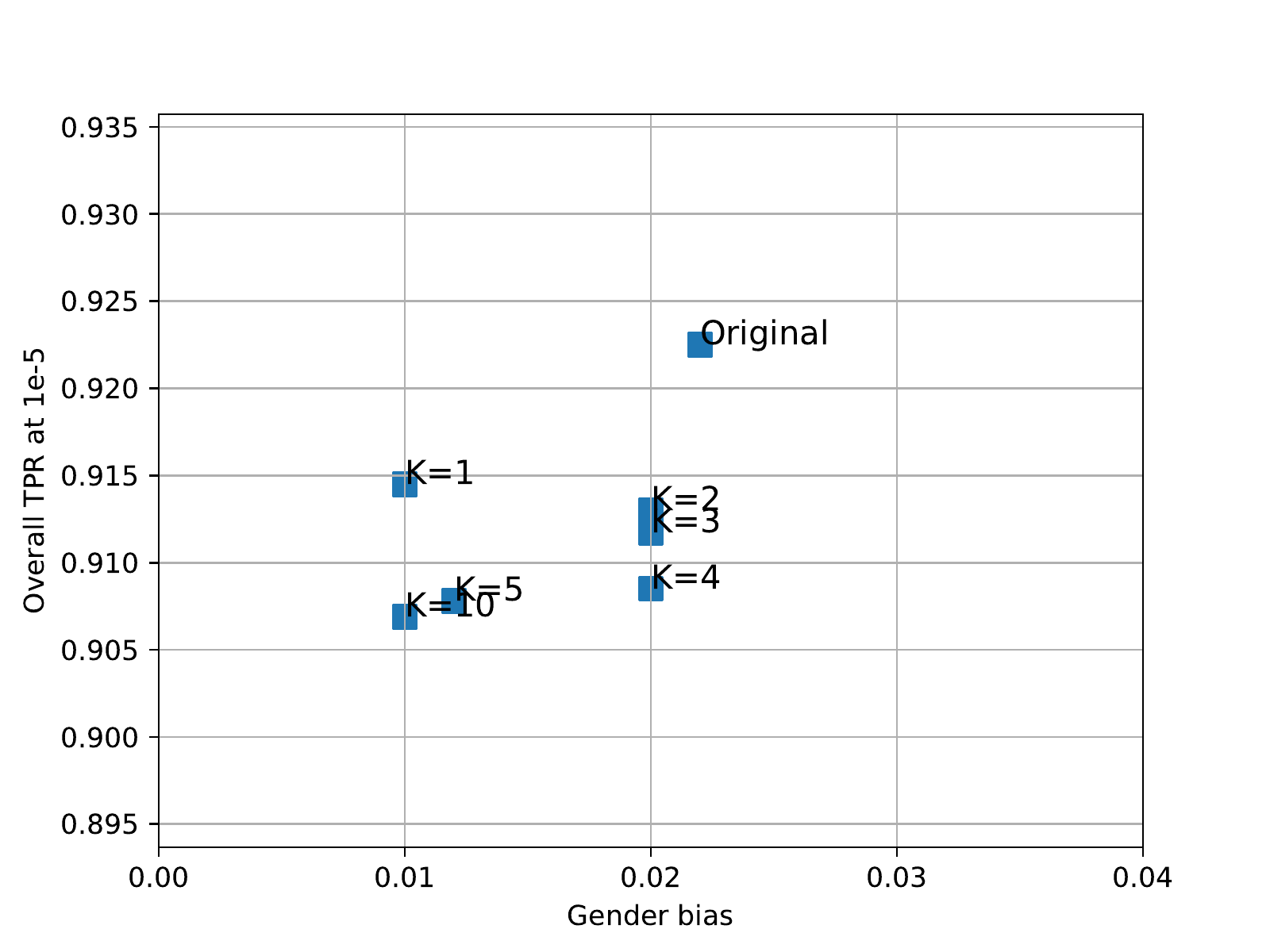}}
\subfloat[Crystalface, $\lambda=1$]{\includegraphics[width=0.25\linewidth]{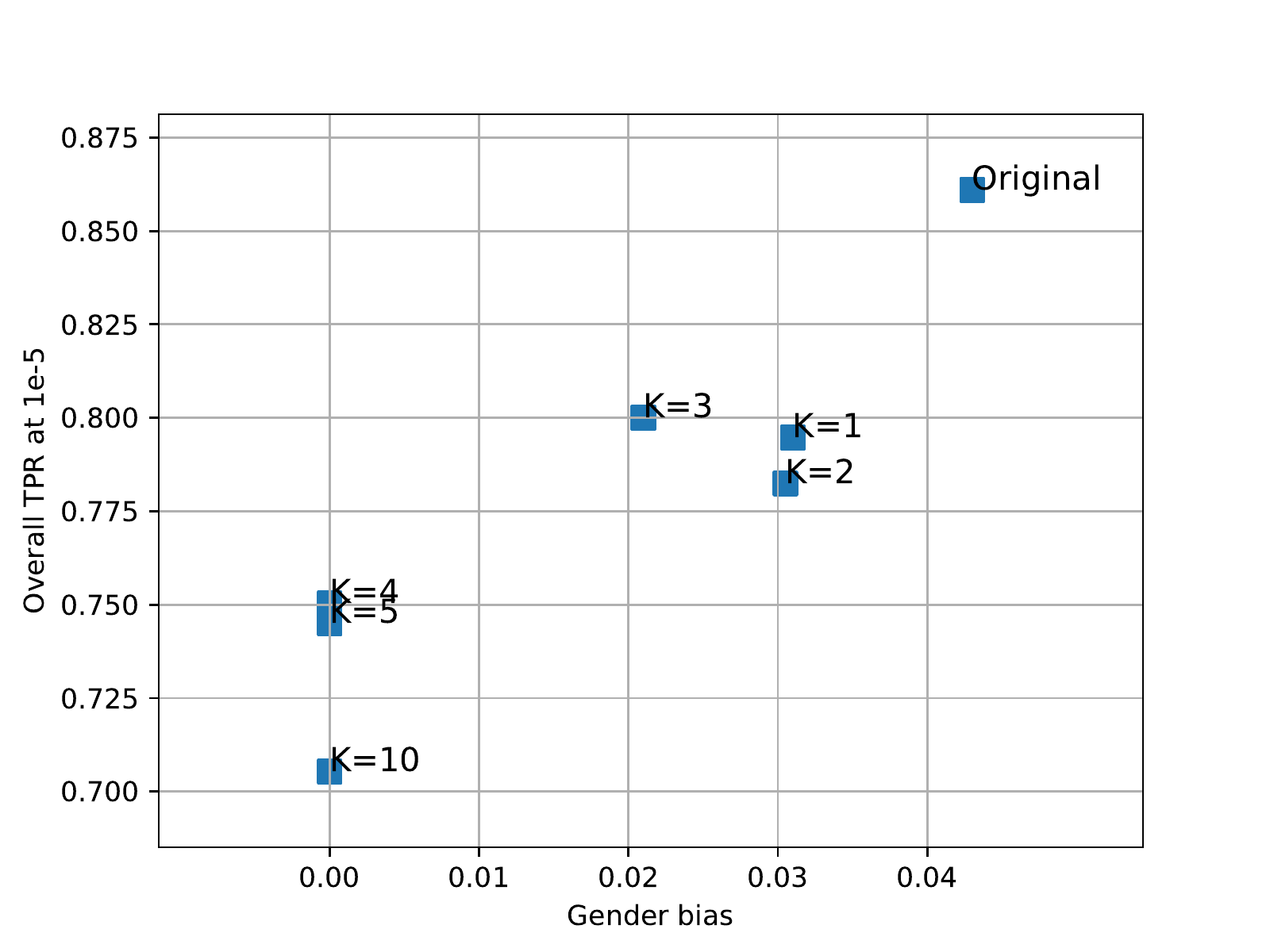}}
\subfloat[Arcface, $K=5$]{\includegraphics[width=0.25\linewidth]{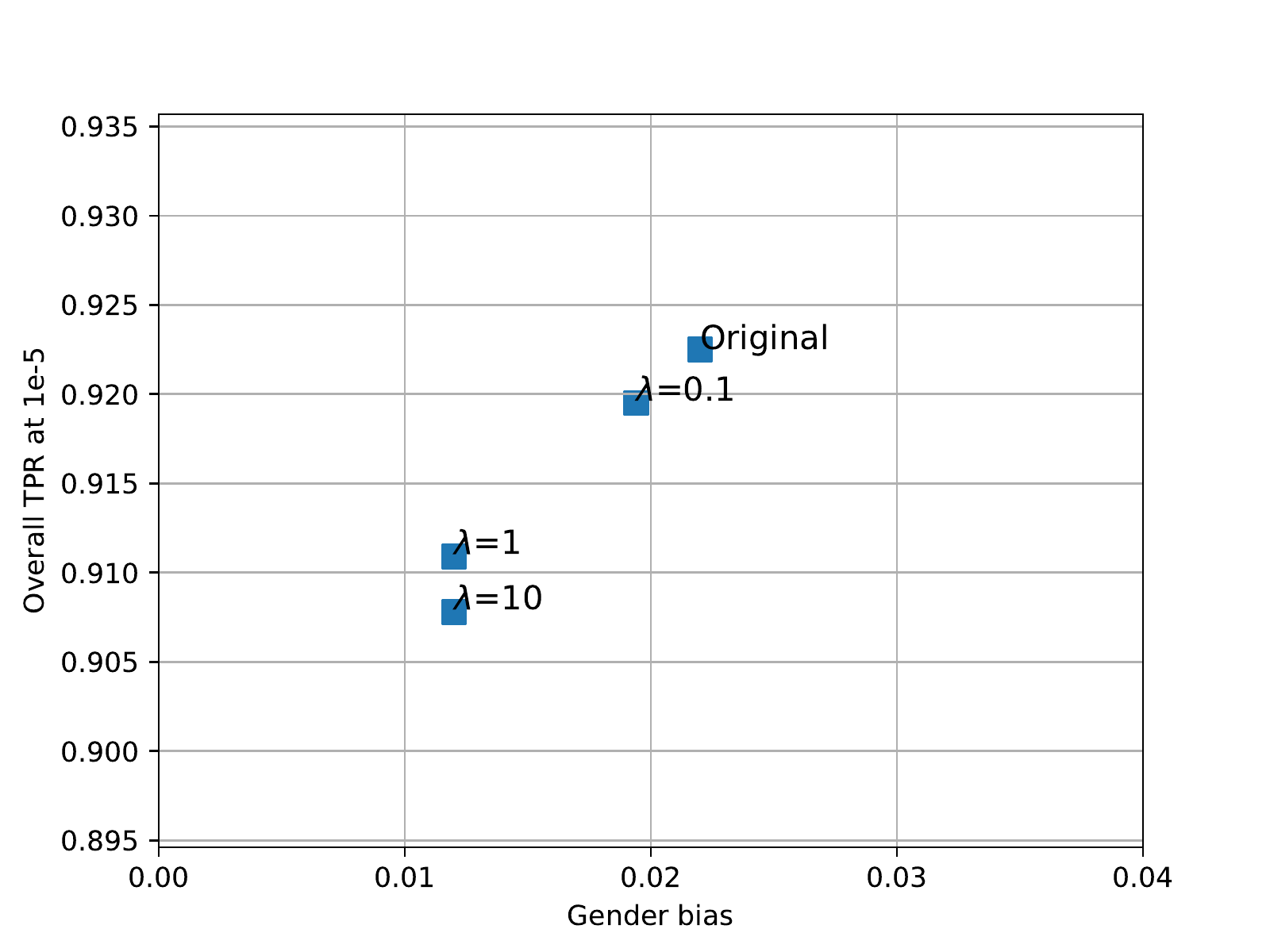}}
\subfloat[Crystalface, $K=5$]{\includegraphics[width=0.25\linewidth]{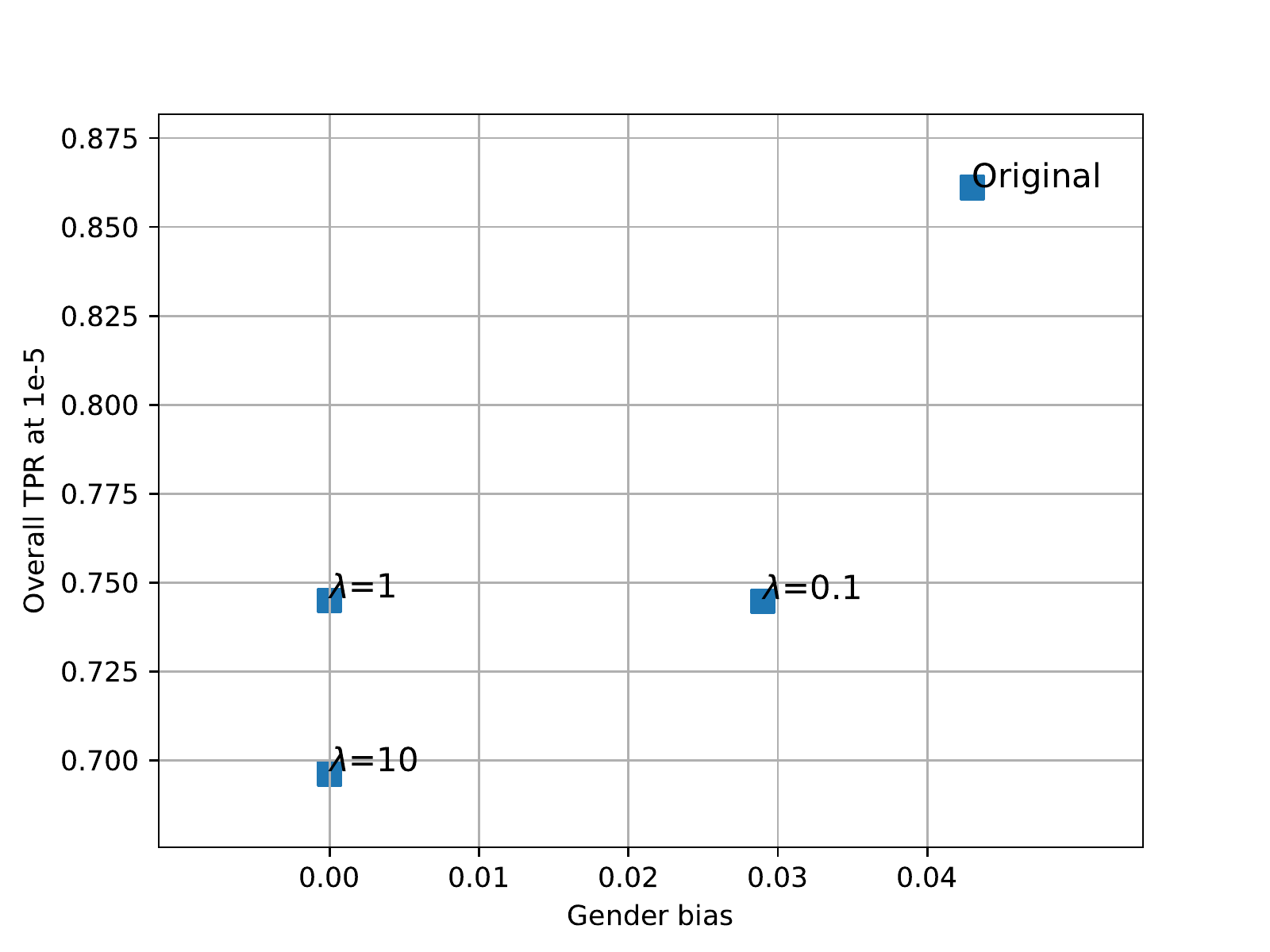}}
\caption{\small \textbf{Ablation experiments:} TPR (at FPR=$10^{-5}$) versus Bias used to train AGENDA on Arcface and Crystalface with different number of gender prediction models $K$ in $E$ (a, b) and with different weights $\lambda$ on $L_{deb}$ (c, d). Evaluation is done on IJBC dataset}
\vspace{-0.3cm}
\label{fig:ablation}
}
\end{figure*}
\vspace{-0.3cm}
\subsection{Ablation study - AGENDA}
Here, we evaluate two hyperparameters used for training the AGENDA framework on Arcface and Crystalface : (a.) the number of gender prediction models $K$ in the ensemble $E$ used to compute $L_{deb}$ (Eq. \ref{eq:ldeb}). (b.) the weight for $L_{deb}$ defined in Eq. \ref{eq:lbr}. We analyze how changing these hyperparameters vary the resultant bias reduction and verification performance at a fixed $\text{FPR}=10^{-5}$ in IJB-C dataset. \\
\textbf{Varying K} :We experiment with $K=1,2,3,4,5$ and $10$. Here, we fix all the other hyperparameters and use the same values specified in Sec. \ref{sec:agendahp}. In Fig. \ref{fig:ablation}(a) and (b), we find that in Arcface, changing $K$ does not have much effect on gender bias or verification TPR at FPR $10^{-5}$. However, for Crystalface, we find that as we increase $K$, the gender bias keeps decreasing which in turn leads to drop in verification performance at FPR $10^{-5}$.  We find that at $K=4$, the bias drops to 0, and as we further keep increasing $K$, the verification performance decreases. \\
\textbf{Varying $\lambda$: } We fix $K=5$ and evaluate $\lambda = 0.1, 1, 10$ for training the AGENDA framework using $f_{in}$. All the other hyperparameters use the same values specified in in Sec. \ref{sec:agendahp}. The results are presented in Fig. \ref{fig:ablation}(c) and (d). For both Arcface and Crystalface, as we keep on increasing the value of $\lambda$, the gender bias keeps generally decreasing and the verification TPR keeps  decreasing.
\begin{figure}
    \centering
    \includegraphics[width=0.4\textwidth]{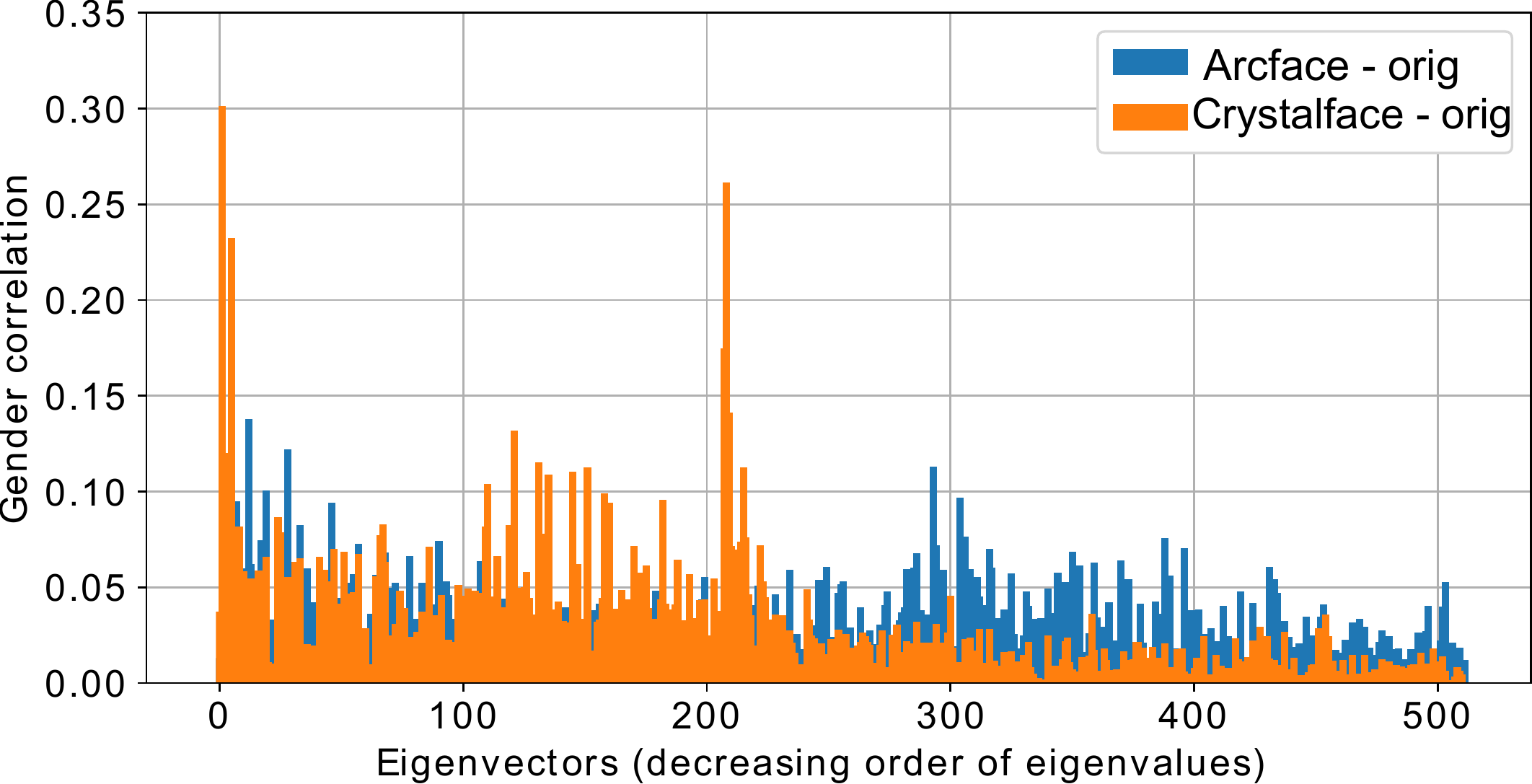}
  \caption{\small Distrubution of gender information in the feature space}
  \vspace{-0.6cm}
  \label{fig:cp}
\end{figure}
\subsection{Analysis of performance drop}
\label{sec:res2}
Gender is an important face attribute that helps deep networks recognize faces. So, minimizing its predictability by using AGENDA is expected to decrease the overall performance as a side-effect. As seen in Fig. \ref{fig:bias_AGENDA}, the verification performance in Crystalface reduces considerably as compared to Arcface, after AGENDA is applied. To understand this behavior, we analyze the distribution of the feature space of both Arcface and Crystalface. For this, we first randomly select 80k images (40k males and females) from the IJB-C dataset, and extract their face descriptors using a given pre-trained network (Arcface or Crystalface). Using PCA, we compute the eigenspace of this set of face descriptors. As done in the implementation of our CorrPCA baseline, we compute the gender correlation of each of the 512 eigenvectors in the eigenspace (Fig. \ref{fig:cp}).

 The top-256 eigenvectors encode more identity information than the bottom ones, since the networks are trained to classify identity. We find that the gender correlation of these identity encoding (top-256) eigenvectors of Crystalface is generally higher than Arcface. This seems to indicate that gender and identity are more entangled in face descriptors from Crystalface than Arcface. Therefore, the drop in verification performance, when the descriptors are gender de-biased is also expected to be more for Crystalface, since the verification performance depends on identity information encoded in the descriptors.
\subsection{Effect of Triplet Probabilistic Embedding}
In \cite{ranjan2019fast}, the face descriptors from Crystalface are not directly used for verification. Instead, the descriptors undergo triplet probabilistic embedding (TPE) \cite{Swami_2016_triplet} for generating a template representation of a given identity. TPE is an embedding learned to generate more discriminative, low-dimensional representations of given input descriptors, that have been shown to achieve better verification results. We apply TPE on the descriptors obtained using Crystalface and find that \textit{TPE improves the overall verification performance, but it also increases bias at all FPRs}. For comparison, we apply TPE after transforming the Crystalface descriptors with AGENDA. From Table \ref{tab:tpeb}, we can infer that the gender bias in the verification results obtained after applying TPE on AGENDA-transformed descriptors is lower than when TPE is applied on original face descriptors of Crystalface. The details of training TPE are provided in the supplementary material.
\begin{table}
\caption{\small IJB-C 1:1 verification results after applying TPE on face descriptors from Crystalface and its AGENDA counterpart}
\scriptsize
\hskip-0.3cm\begin{tabular}{cccc|ccc|ccc|ccc}
\toprule
FPR & \multicolumn{3}{c|}{$10^{-6}$}& \multicolumn{3}{c|}{$10^{-5}$} & \multicolumn{3}{c|}{$10^{-4}$} & \multicolumn{3}{c}{$10^{-3}$}\\
\midrule
Method& TPR\textsubscript{m} & TPR\textsubscript{f}& Bias & TPR\textsubscript{m} & TPR\textsubscript{f}& Bias & TPR\textsubscript{m} & TPR\textsubscript{f}& Bias & TPR\textsubscript{m} & TPR\textsubscript{f}& Bias\\
\midrule
Orig. + TPE& 0.80&0.69& 0.11 & 0.88 & 0.84 & 0.04 &0.93 & 0.89 & 0.04 &0.96&0.94&0.02 \\
AGENDA +TPE & 0.57&0.51 & \textbf{0.06}&0.75& 0.73& \textbf{0.02} & 0.88 & 0.85 & \textbf{0.03} &0.93&0.91&0.02\\
\bottomrule
\end{tabular}
\vspace{-0.5cm}
\label{tab:tpeb}
\end{table}
\section{Conclusion}
Implicit encoding of gender in face descriptors during training may lead to privacy leakage as such descriptors can be trained to classify face gender. Moreover, the expression of gender also appears to contribute to gender bias when such descriptors are used for face verification.  We address the issue of annonymizing the gender of face descriptors. Our initial experiments show that face descriptors showing lower gender predictability generally demonstrate lower gender bias in face verification. Motivated by this finding and the need for anonymizing the gender of face representations, we propose a framework - AGENDA, that adversarially reduces gender information from face descriptors, while training them to classify identities. The results of our experiments with AGENDA and CorrPCA provide evidence in support of our hypothesis that reduction of gender predictability in face descriptors will decrease gender bias in face verification. AGENDA significantly outperforms our PCA-based baseline in terms of bias reduction. The efficacy of AGENDA is evaluated on face descriptors from two SOTA recognition networks. However, gender is an important facial attribute, and we find that reducing gender information leads to slight decrease in verification accuracy. In the near future, we intend to apply and modify CorrPCA and AGENDA to reduce the information of other attributes like age and race in face recognition features; and apply a combination of several de-biasing losses to reduce the strength of multiple attributes simultaneously.
\section*{Acknowledgement}
The authors would like to thank P. Jonathon Phillips, Ankan Bansal, Aniket Roy and Rajeev Ranjan for their helpful suggestions. This research is based upon work supported by the Office of the Director of National Intelligence (ODNI), Intelligence Advanced Research Projects Activity (IARPA), via IARPA R\&D Contract No. 2019-022600002. The views and conclusions contained herein are those of the authors and should not be interpreted as necessarily representing the official policies or endorsements, either expressed or implied, of the ODNI, IARPA, or the U.S. Government. The U.S. Government is authorized to reproduce and distribute reprints for Governmental purposes notwithstanding any copyright annotation thereon.
\bibstyle{aaai21}
\bibliography{egbib}

\begin{thebibliography}{45}
\providecommand{\natexlab}[1]{#1}
\providecommand{\url}[1]{\texttt{#1}}
\providecommand{\urlprefix}{URL }
\expandafter\ifx\csname urlstyle\endcsname\relax
  \providecommand{\doi}[1]{doi:\discretionary{}{}{}#1}\else
  \providecommand{\doi}{doi:\discretionary{}{}{}\begingroup
  \urlstyle{rm}\Url}\fi

\bibitem[{Albiero and Bowyer(2020)}]{albiero2020face}
Albiero, V.; and Bowyer, K.~W. 2020.
\newblock Is Face Recognition Sexist? No, Gendered Hairstyles and Biology Are.
\newblock \emph{arXiv preprint arXiv:2008.06989} .

\bibitem[{Albiero et~al.(2020)Albiero, KS, Vangara, Zhang, King, and
  Bowyer}]{albiero2020analysis}
Albiero, V.; KS, K.; Vangara, K.; Zhang, K.; King, M.~C.; and Bowyer, K.~W.
  2020.
\newblock Analysis of gender inequality in face recognition accuracy.
\newblock In \emph{Proceedings of the IEEE Winter Conference on Applications of
  Computer Vision Workshops}, 81--89.

\bibitem[{Albiero, Zhang, and Bowyer(2020)}]{albiero2020does}
Albiero, V.; Zhang, K.; and Bowyer, K.~W. 2020.
\newblock How Does Gender Balance In Training Data Affect Face Recognition
  Accuracy?
\newblock \emph{arXiv preprint arXiv:2002.02934} .

\bibitem[{Alvi, Zisserman, and Nell{\aa}ker(2018)}]{alvi2018turning}
Alvi, M.; Zisserman, A.; and Nell{\aa}ker, C. 2018.
\newblock Turning a blind eye: Explicit removal of biases and variation from
  deep neural network embeddings.
\newblock In \emph{Proceedings of the European Conference on Computer Vision
  (ECCV)}, 0--0.

\bibitem[{Amini et~al.(2019)Amini, Soleimany, Schwarting, Bhatia, and
  Rus}]{amini2019uncovering}
Amini, A.; Soleimany, A.~P.; Schwarting, W.; Bhatia, S.~N.; and Rus, D. 2019.
\newblock Uncovering and mitigating algorithmic bias through learned latent
  structure.
\newblock In \emph{Proceedings of the 2019 AAAI/ACM Conference on AI, Ethics,
  and Society}, 289--295.

\bibitem[{Bansal et~al.(2017{\natexlab{a}})Bansal, Castillo, Ranjan, and
  Chellappa}]{bansal2017s}
Bansal, A.; Castillo, C.~D.; Ranjan, R.; and Chellappa, R. 2017{\natexlab{a}}.
\newblock The do's and don'ts for {CNN}-based face verification.
\newblock In \emph{Proceedings of the IEEE International Conference on Computer
  Vision}, 2545--2554.

\bibitem[{Bansal et~al.(2017{\natexlab{b}})Bansal, Nanduri, Castillo, Ranjan,
  and Chellappa}]{bansal2017umdfaces}
Bansal, A.; Nanduri, A.; Castillo, C.~D.; Ranjan, R.; and Chellappa, R.
  2017{\natexlab{b}}.
\newblock Umdfaces: An annotated face dataset for training deep networks.
\newblock In \emph{2017 IEEE International Joint Conference on Biometrics
  (IJCB)}, 464--473. IEEE.

\bibitem[{Bansal et~al.(2018)Bansal, Ranjan, Castillo, and
  Chellappa}]{bansal2018deep}
Bansal, A.; Ranjan, R.; Castillo, C.~D.; and Chellappa, R. 2018.
\newblock Deep Features for Recognizing Disguised Faces in the Wild.
\newblock In \emph{2018 IEEE/CVF Conference on Computer Vision and Pattern
  Recognition Workshops (CVPRW)}, 10--106. IEEE.

\bibitem[{Buolamwini and Gebru(2018)}]{buolamwini2018gender}
Buolamwini, J.; and Gebru, T. 2018.
\newblock Gender shades: Intersectional accuracy disparities in commercial
  gender classification.
\newblock In \emph{Conference on fairness, accountability and transparency},
  77--91.

\bibitem[{Cavazos et~al.(2019)Cavazos, Phillips, Castillo, and
  O'Toole}]{cavazos2019accuracy}
Cavazos, J.~G.; Phillips, P.~J.; Castillo, C.~D.; and O'Toole, A.~J. 2019.
\newblock Accuracy comparison across face recognition algorithms: Where are we
  on measuring race bias?

\bibitem[{Cook et~al.(2019)Cook, Howard, Sirotin, and Tipton}]{cook2019fixed}
Cook, C.; Howard, J.; Sirotin, Y.; and Tipton, J. 2019.
\newblock Fixed and Varying Effects of Demographic Factors on the Performance
  of Eleven Commercial Facial Recognition Systems.
\newblock \emph{IEEE Transactions on Biometrics, Behavior, and Identity
  Science} 40(1).

\bibitem[{Deng et~al.(2019)Deng, Guo, Niannan, and Zafeiriou}]{deng2018arcface}
Deng, J.; Guo, J.; Niannan, X.; and Zafeiriou, S. 2019.
\newblock ArcFace: Additive Angular Margin Loss for Deep Face Recognition.
\newblock In \emph{CVPR}.

\bibitem[{Dhar et~al.(2020)Dhar, Bansal, Castillo, Gleason, Phillips, and
  Chellappa}]{dhar2019attributes}
Dhar, P.; Bansal, A.; Castillo, C.~D.; Gleason, J.; Phillips, P.~J.; and
  Chellappa, R. 2020.
\newblock How are attributes expressed in face {DCNNs?}
\newblock \emph{To appear in 15th IEEE Intl. Conf. Automatic Face and Gesture
  Recognition, arXiv preprint arXiv:1910.05657} .

\bibitem[{Drozdowski et~al.(2020)Drozdowski, Rathgeb, Dantcheva, Damer, and
  Busch}]{drozdowski2020demographic}
Drozdowski, P.; Rathgeb, C.; Dantcheva, A.; Damer, N.; and Busch, C. 2020.
\newblock Demographic Bias in Biometrics: A Survey on an Emerging Challenge.
\newblock \emph{arXiv preprint arXiv:2003.02488} .

\bibitem[{Fu, Guo, and Huang(2010)}]{fu2010age}
Fu, Y.; Guo, G.; and Huang, T. 2010.
\newblock Age synthesis and estimation via faces: A survey.
\newblock \emph{IEEE Transactions on Pattern Analysis and Machine Intelligence}
  32(11): 1955--1976.

\bibitem[{Furl, Phillips, and O'Toole(2002)}]{article}
Furl, N.; Phillips, P.~J.; and O'Toole, A. 2002.
\newblock Face recognition algorithms and the other-race effect: Computational
  mechanisms for a developmental contact hypothesis.
\newblock \emph{Cognitive Science} 26: 797--815.
\newblock \doi{10.1016/S0364-0213(02)00084-8}.

\bibitem[{Georgopoulos, Panagakis, and
  Pantic(2020)}]{georgopoulos2020investigating}
Georgopoulos, M.; Panagakis, Y.; and Pantic, M. 2020.
\newblock Investigating Bias in Deep Face Analysis: The KANFace Dataset and
  Empirical Study.
\newblock \emph{arXiv preprint arXiv:2005.07302} .

\bibitem[{Grother, Ngan, and Hanaoka(2019)}]{grother2019face}
Grother, P.; Ngan, M.; and Hanaoka, K. 2019.
\newblock Face Recognition Vendor Test ({FRVT}) Part 3: Demographic Effects.
\newblock \emph{National Institute of Standards and Technology} .

\bibitem[{Guo et~al.(2016)Guo, Zhang, Hu, He, and Gao}]{guo2016ms}
Guo, Y.; Zhang, L.; Hu, Y.; He, X.; and Gao, J. 2016.
\newblock MS-Celeb-1M: A Dataset and Benchmark for Large-Scale Face
  Recognition.
\newblock In \emph{European Conference on Computer Vision}, 87--102. Springer.

\bibitem[{He et~al.(2015)He, Zhang, Ren, and Sun}]{he2015delving}
He, K.; Zhang, X.; Ren, S.; and Sun, J. 2015.
\newblock Delving deep into rectifiers: Surpassing human-level performance on
  imagenet classification.
\newblock In \emph{Proceedings of the IEEE international conference on computer
  vision}, 1026--1034.

\bibitem[{Hill et~al.(2019)Hill, Parde, Castillo, Colon, Ranjan, Chen, Blanz,
  and O’Toole}]{hill2019deep}
Hill, M.~Q.; Parde, C.~J.; Castillo, C.~D.; Colon, Y.~I.; Ranjan, R.; Chen,
  J.-C.; Blanz, V.; and O’Toole, A.~J. 2019.
\newblock Deep convolutional neural networks in the face of caricature.
\newblock \emph{Nature Machine Intelligence} 1(11): 522--529.

\bibitem[{Klambauer et~al.(2017)Klambauer, Unterthiner, Mayr, and
  Hochreiter}]{klambauer2017self}
Klambauer, G.; Unterthiner, T.; Mayr, A.; and Hochreiter, S. 2017.
\newblock Self-normalizing neural networks.
\newblock In \emph{Advances in neural information processing systems},
  971--980.

\bibitem[{Klare et~al.(2012)Klare, Burge, Klontz, Bruegge, and
  Jain}]{klare2012face}
Klare, B.~F.; Burge, M.~J.; Klontz, J.~C.; Bruegge, R. W.~V.; and Jain, A.~K.
  2012.
\newblock Face recognition performance: Role of demographic information.
\newblock \emph{IEEE Transactions on Information Forensics and Security} 7(6):
  1789--1801.

\bibitem[{Krishnapriya et~al.(2020)Krishnapriya, Albiero, Vangara, King, and
  Bowyer}]{krishnapriya2020issues}
Krishnapriya, K.; Albiero, V.; Vangara, K.; King, M.~C.; and Bowyer, K.~W.
  2020.
\newblock Issues Related to Face Recognition Accuracy Varying Based on Race and
  Skin Tone.
\newblock \emph{IEEE Transactions on Technology and Society} 1(1): 8--20.

\bibitem[{Li et~al.(2019)Li, Guo, Yang, and Chen}]{li2019deepobfuscator}
Li, A.; Guo, J.; Yang, H.; and Chen, Y. 2019.
\newblock Deepobfuscator: Adversarial training framework for privacy-preserving
  image classification.
\newblock \emph{arXiv preprint arXiv:1909.04126} .

\bibitem[{Liu et~al.(2015)Liu, Luo, Wang, and Tang}]{liu2015faceattributes}
Liu, Z.; Luo, P.; Wang, X.; and Tang, X. 2015.
\newblock Deep Learning Face Attributes in the Wild.
\newblock In \emph{Proceedings of International Conference on Computer Vision
  (ICCV)}.

\bibitem[{Lu, Jain et~al.(2004)}]{lu2004ethnicity}
Lu, X.; Jain, A.~K.; et~al. 2004.
\newblock Ethnicity identification from face images.
\newblock In \emph{Proceedings of SPIE}, volume 5404, 114--123.

\bibitem[{Makinen and Raisamo(2008)}]{makinen2008evaluation}
Makinen, E.; and Raisamo, R. 2008.
\newblock Evaluation of gender classification methods with automatically
  detected and aligned faces.
\newblock \emph{IEEE transactions on pattern analysis and machine intelligence}
  30(3): 541--547.

\bibitem[{Maze et~al.(2018)Maze, Adams, Duncan, Kalka, Miller, Otto, Jain,
  Niggel, Anderson, Cheney et~al.}]{maze2018iarpa}
Maze, B.; Adams, J.; Duncan, J.~A.; Kalka, N.; Miller, T.; Otto, C.; Jain,
  A.~K.; Niggel, W.~T.; Anderson, J.; Cheney, J.; et~al. 2018.
\newblock {IARPA} janus benchmark-c: Face dataset and protocol.
\newblock In \emph{2018 International Conference on Biometrics (ICB)},
  158--165. IEEE.

\bibitem[{Mirjalili, Raschka, and Ross(2018)}]{mirjalili2018gender}
Mirjalili, V.; Raschka, S.; and Ross, A. 2018.
\newblock Gender privacy: An ensemble of semi adversarial networks for
  confounding arbitrary gender classifiers.
\newblock In \emph{2018 IEEE 9th International Conference on Biometrics Theory,
  Applications and Systems (BTAS)}, 1--10. IEEE.

\bibitem[{Mirjalili and Ross(2017)}]{mirjalili2017soft}
Mirjalili, V.; and Ross, A. 2017.
\newblock Soft biometric privacy: Retaining biometric utility of face images
  while perturbing gender.
\newblock In \emph{2017 IEEE International joint conference on biometrics
  (IJCB)}, 564--573. IEEE.

\bibitem[{Nagpal et~al.(2019)Nagpal, Singh, Singh, Vatsa, and
  Ratha}]{nagpal2019deep}
Nagpal, S.; Singh, M.; Singh, R.; Vatsa, M.; and Ratha, N. 2019.
\newblock Deep Learning for Face Recognition: Pride or Prejudiced?
\newblock \emph{arXiv preprint arXiv:1904.01219} .

\bibitem[{Othman and Ross(2014)}]{othman2014privacy}
Othman, A.; and Ross, A. 2014.
\newblock Privacy of facial soft biometrics: Suppressing gender but retaining
  identity.
\newblock In \emph{European Conference on Computer Vision}, 682--696. Springer.

\bibitem[{Ranjan et~al.(2019)Ranjan, Bansal, Zheng, Xu, Gleason, Lu, Nanduri,
  Chen, Castillo, and Chellappa}]{ranjan2019fast}
Ranjan, R.; Bansal, A.; Zheng, J.; Xu, H.; Gleason, J.; Lu, B.; Nanduri, A.;
  Chen, J.-C.; Castillo, C.~D.; and Chellappa, R. 2019.
\newblock A fast and accurate system for face detection, identification, and
  verification.
\newblock \emph{IEEE Transactions on Biometrics, Behavior, and Identity
  Science} 1(2): 82--96.

\bibitem[{Ranjan et~al.(2017)Ranjan, Sankaranarayanan, Castillo, and
  Chellappa}]{ranjan2017all}
Ranjan, R.; Sankaranarayanan, S.; Castillo, C.~D.; and Chellappa, R. 2017.
\newblock An all-in-one convolutional neural network for face analysis.
\newblock In \emph{2017 12th IEEE International Conference on Automatic Face \&
  Gesture Recognition (FG 2017)}, 17--24. IEEE.

\bibitem[{Sankaranarayanan et~al.(2016)Sankaranarayanan, Alavi, Castillo, and
  Chellappa}]{Swami_2016_triplet}
Sankaranarayanan, S.; Alavi, A.; Castillo, C.~D.; and Chellappa, R. 2016.
\newblock Triplet Probabilistic Embedding for Face Verification and Clustering.
\newblock In \emph{2016 IEEE 8th International Conference on Biometrics Theory,
  Applications and Systems (BTAS)}.

\bibitem[{Sattigeri et~al.(2018)Sattigeri, Hoffman, Chenthamarakshan, and
  Varshney}]{sattigeri2018fairness}
Sattigeri, P.; Hoffman, S.~C.; Chenthamarakshan, V.; and Varshney, K.~R. 2018.
\newblock Fairness {GAN}.
\newblock \emph{arXiv preprint arXiv:1805.09910} .

\bibitem[{Schroff, Kalenichenko, and Philbin(2015)}]{schroff2015facenet}
Schroff, F.; Kalenichenko, D.; and Philbin, J. 2015.
\newblock Facenet: A unified embedding for face recognition and clustering.
\newblock In \emph{Proceedings of the IEEE Conference on Computer Vision and
  Pattern Recognition}, 815--823.

\bibitem[{Taigman et~al.(2014)Taigman, Yang, Ranzato, and
  Wolf}]{taigman2014deepface}
Taigman, Y.; Yang, M.; Ranzato, M.; and Wolf, L. 2014.
\newblock Deepface: Closing the gap to human-level performance in face
  verification.
\newblock In \emph{Proceedings of the IEEE Conference on Computer Vision and
  Pattern Recognition}, 1701--1708.

\bibitem[{Turk and Pentland(1991)}]{turk1991face}
Turk, M.; and Pentland, A. 1991.
\newblock Face recognition using eigenfaces.
\newblock In \emph{Proceedings. 1991 IEEE computer society conference on
  computer vision and pattern recognition}, 586--587.

\bibitem[{Vangara et~al.(2019)Vangara, King, Albiero, Bowyer
  et~al.}]{vangara2019characterizing}
Vangara, K.; King, M.~C.; Albiero, V.; Bowyer, K.; et~al. 2019.
\newblock Characterizing the Variability in Face Recognition Accuracy Relative
  to Race.
\newblock In \emph{Proceedings of the IEEE Conference on Computer Vision and
  Pattern Recognition Workshops}, 0--0.

\bibitem[{Wang and Deng(2020)}]{wang2020mitigating}
Wang, M.; and Deng, W. 2020.
\newblock Mitigating Bias in Face Recognition Using Skewness-Aware
  Reinforcement Learning.
\newblock In \emph{Proceedings of the IEEE/CVF Conference on Computer Vision
  and Pattern Recognition}, 9322--9331.

\bibitem[{Wang et~al.(2019{\natexlab{a}})Wang, Deng, Hu, Tao, and
  Huang}]{wang2019racial}
Wang, M.; Deng, W.; Hu, J.; Tao, X.; and Huang, Y. 2019{\natexlab{a}}.
\newblock Racial Faces in the Wild: Reducing Racial Bias by Information
  Maximization Adaptation Network.
\newblock In \emph{Proceedings of the IEEE International Conference on Computer
  Vision}, 692--702.

\bibitem[{Wang et~al.(2019{\natexlab{b}})Wang, Zhao, Yatskar, Chang, and
  Ordonez}]{wang2019balanced}
Wang, T.; Zhao, J.; Yatskar, M.; Chang, K.-W.; and Ordonez, V.
  2019{\natexlab{b}}.
\newblock Balanced datasets are not enough: Estimating and mitigating gender
  bias in deep image representations.
\newblock In \emph{Proceedings of the IEEE International Conference on Computer
  Vision}, 5310--5319.

\bibitem[{Wu et~al.(2018)Wu, Wang, Wang, and Jin}]{wu2018towards}
Wu, Z.; Wang, Z.; Wang, Z.; and Jin, H. 2018.
\newblock Towards privacy-preserving visual recognition via adversarial
  training: A pilot study.
\newblock In \emph{Proceedings of the European Conference on Computer Vision
  (ECCV)}, 606--624.

\end{thebibliography}
\appendix
\renewcommand\thesection{\Alph{section}}
\renewcommand\thesubsection{\thesection.\arabic{subsection}}
\section{Motivation for training one model in $E$}
In Stage 4 of Sec. 4.2, we heuristically choose a model $E_k$ in the ensemble $E$ and re-train it to classify gender using $f_{out}$ (from model $M$) as input. This is also described in Step 26 and 28 of Algorithm 1. In this section we describe the motivation for that decision.

The main idea here is to ensure that the ensemble of models predicting gender is diverse. The desire for diversity relates to the `the $\forall$ challenge' described by \cite{wu2018towards} which, in the context of our work, implies that $f_{out}$ must be transformed such that gender cannot be extracted using any model. The overarching idea for holding all but one model in ensemble $E$ constant within one training episode is to restrict $M$ from learning to embed gender information in successively larger regions of the descriptor space. The remainder of this section elaborates on what we mean by this.

Treating $f_{out}$ and $f_{in}$ as random variables, let $p(f_{out} | male)$ be the likelihood of male face descriptors and $p(f_{out} | female)$ be the likelihood of female face descriptors. In order to guarantee that the mutual information between $f_{out}$ and gender is zero, $p(f_{out} | male)$ and $p(f_{out} | female)$ must be equal almost everywhere for $f_{out} \in M(F_{in})$ where $F_{in}$ is the support of $f_{in}$ (i.e. the set of values which $p(f_{in}) \neq 0$).

For the purpose of this motivation we suppose our objective is to minimize the measure of set $\mathcal{G} = \left\{f_{out} = M(f_{in})~|~p(f_{out} | male) \neq p(f_{out} | female) \right\}$, thus ensuring that $f_{out}$ contains no information about gender.

Further, let $\mathcal{E}$ represent the region of $\mathbb{R}^{256}$ where any model within ensemble $E$ predicts the probability of male and female to be non-equal. Supposing that $E$ is composed entirely of optimal classifiers, then $\mathcal{E}$ would be equal to $\mathcal{G}$.

Suppose that $\mathcal{E}_t$ and $\mathcal{G}_t$ are $\mathcal{G}$ and $\mathcal{E}$ after the $t$th episode of training from Algorithm 1. As an alternative, consider a version of the algorithm where every model in $E$ is trained at each episode, rather than training only one at a time. Since $M$ is trained to confuse $E$ using $L_{deb}$, a plausible strategy is that $M$ will be updated so that $\mathcal{G}_{t+1} \cap \mathcal{E}_t = \emptyset$. This strategy minimizes the debiasing loss $L_{deb}$, since each model in ensemble $E$ is successfully fooled, but does not necessarily reduce the measure of $\mathcal{G}$. This strategy is also, in some sense, easier than the true objective of minimizing $\mathcal{G}$, since $M$ does not need to disentangle and remove the representation of gender from $f_{out}$. Instead, $M$ could modify the representation so that the same information is present, but represented in an alternative subset of the descriptor space.

The strategy of holding all but one of the models in $E$ constant is an attempt to mitigate this potential issue. Again, assuming each model in $E$ to be optimal, then by training only a single model within $E$ at each episode, $\mathcal{E}_t = \mathcal{G}_{t-(K-1)} \cup \ldots \cup \mathcal{G}_t$, thus preventing $\mathcal{G}$ from oscillating between two disjoint regions of feature space. Furthermore, in an ideal scenario with unlimited training time and memory, $K$ should be equal to the total number of training episodes. This would ensure that $M$ can never encode gender information the same way twice, which would successively make it more likely to actually reduce the measure of $\mathcal{G}$. In practice, however, the total memory and training time is limited, so $K$ is chosen to be as large as practical.








\section{Details for training TPE}
To learn a triplet probabilistic embedding $W_{cf}$, we use the descriptors from Crystalface (extracted for UMD-Faces \cite{bansal2017umdfaces} dataset). This embedding $W_{cf} \in \mathbb{R}^{512\times128}$ is then used to transform the 512 dimensional IJB-C \cite{maze2018iarpa} descriptors (extracted using Crystalface) to obtain 128-dimensional face descriptors, which are used for 1:1 face verification. The results of this experiment are provided in `Orig +TPE' in Table 4 of Sec. 5.8. We perform the same experiment with the AGENDA-transformed descriptors of Crystalface, where a new TPE matrix  $W'_{cf} \in \mathbb{R}^{256\times128}$ is learned and used to transform the IJB-C descriptors before performing 1:1 verification. The results of this experiment are provided in `AGENDA +TPE' in Table 4 of Sec. 5.8. We find that the results obtained after applying TPE on AGENDA-transformed  features have relatively lower bias.\\

For training both, $W_{cf}$ and $W'_{cf}$, we use a fixed learning rate of $2.5 \times 10^{-3}$ and a batch size of 32. The training for computing such a matrix using the descriptors from Crystalface (or its AGENDA counterpart) generally converges after 10k iterations.  For a given set of descriptors, we compute its TPE matrix ten times and finally compute the  average of the resulting matrices. We use this matrix to transform the test descriptors. More details about TPE are provided in \cite{Swami_2016_triplet}.
\end{document}